\documentclass[10pt,twocolumn,letterpaper]{article}

\usepackage{pgfplots}
\pgfplotsset{compat=1.16}
\usepackage[hypcap=false]{caption}
\usepackage[pagenumbers]{cvpr} 

\definecolor{cvprblue}{rgb}{0.21,0.49,0.74}
\usepackage{hyperref}

\def\paperID{7586} 
\def\confName{CVPR}
\def\confYear{2025}

\title{SliderSpace: Decomposing the Visual Capabilities of Diffusion Models}

\author{Rohit Gandikota$^{1*}$ \hspace{.7em} Zongze Wu$^{2}$ \hspace{.7em} Richard Zhang$^{2}$\hspace{.7em} David Bau$^{1}$ \hspace{.7em} Eli Shechtman$^{2}$ \hspace{.7em}  Nick Kolkin$^{2}$ \vspace{3pt} \\ 
$^{1}$Northeastern University  \quad $^{2}$Adobe Research}

\begin{document}

\twocolumn[{
\renewcommand\twocolumn[1][]{#1}
\maketitle
\includegraphics[width=\textwidth]{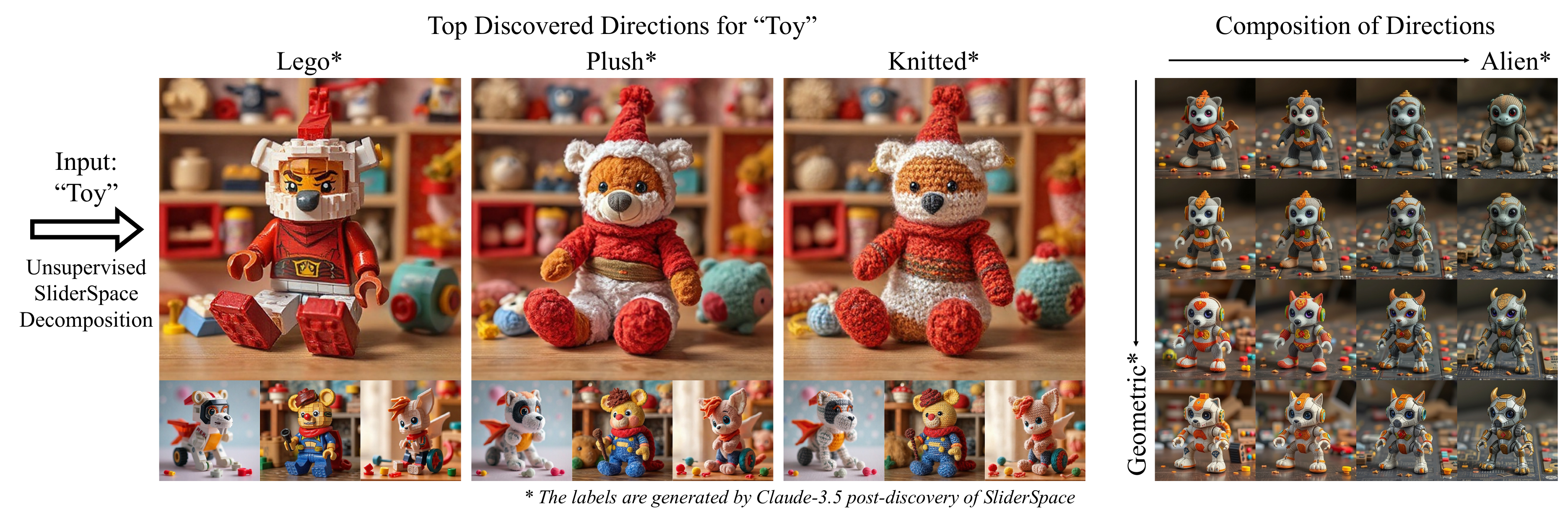}
\captionof{figure}{Given a prompt, SliderSpace identifies the principal directions of the visual capabilites of a diffusion model by decomposing the image distribution over the prompt. By manipulating these directions as sliders, users can control and combine them to explore the creative possibilities of the model. We visualize the top directions discovered by SliderSpace within Flux Schnell~\cite{flux} for the concept ``Toy''.\vspace{2em}}
\label{fig:intro}
\vspace{-2mm}
}]

\begin{abstract}

We present SliderSpace, a framework for automatically decomposing the visual capabilities of diffusion models into controllable and human-understandable directions. Unlike existing control methods that require a user to specify attributes for each edit direction individually, SliderSpace discovers multiple interpretable and diverse directions simultaneously from a single text prompt. Each direction is trained as a low-rank adaptor, enabling compositional control and the discovery of surprising possibilities in the model's latent space. Through extensive experiments on state-of-the-art diffusion models, we demonstrate SliderSpace's effectiveness across three applications: concept decomposition, artistic style exploration, and diversity enhancement. Our quantitative evaluation shows that SliderSpace-discovered directions decompose the visual structure of model's knowledge effectively, offering insights into the latent capabilities encoded within diffusion models. User studies further validate that our method produces more diverse and useful variations compared to baselines. 
Our code, data and trained weights are available at \href{https://sliderspace.baulab.info/}{\textcolor[rgb]{0.21,0.49,0.74}{sliderspace.baulab.info}}

\end{abstract}   

\section{Introduction}
\label{sec:intro}

Text-to-image diffusion models are capable of generating remarkable visual variations from a single prompt through different random initializations. However, this vast creative potential remains largely opaque to users---while we can generate diverse images, we lack understanding of the underlying structure of these variations. This presents a fundamental challenge: how can we discover and expose the latent visual capabilities encoded within these models?

\let\thefootnote\relax \footnote{$^{*}$Correspondence to \texttt{gandikota.ro@northeastern.edu}}

The challenge touches on a key limitation in how we interact with diffusion models today. Current control methods require users to explicitly specify their desired edits in advance through prompts~\cite{gandikota2023concept}, reference images~\cite{zhang2023addingconditionalcontroltexttoimage, chen2024trainingfreeregionalpromptingdiffusion, ruiz2022dreambooth,kumari2022customdiffusion, Ryu_lora, hu2021lora}, or attribute vectors~\cite{ye2023ipadaptertextcompatibleimage, hertz2024stylealignedimagegeneration, li2023photomaker, shi2024instantbooth,parmar2023zero,hertz2022prompt}. That contrasts sharply with natural human creative workflows, where artists dynamically explore creative ideas and jointly refine them toward meaningful artistic outcomes~\cite{hoffmann2016modeling}. The need for pre-specified controls creates a barrier between users and the full creative potential of these models.

Interestingly, earlier generative models like GANs~\cite{gans,karras2019style,brock2018large} naturally developed more interpretable internal structures. Their compact latent spaces often exhibited emergent disentanglement~\cite{harkonen2020ganspace,radford2015unsupervised, wu2021stylespace, shen2020interfacegan}, enabling continuous and compositional control over generated images. Users could explore these spaces to discover interesting variations that would be difficult to describe in words~\cite{wu2021stylespace}, then combine them to achieve their creative goals~\cite{grabe2022towards}.

Diffusion models have largely superseded GANs in conditional image synthesis~\cite{dhariwal2021diffusion}, achieving greater diversity through much higher-dimensional latents. And yet an understanding of the underlying structure of these larger latent spaces has remained elusive. In this work, we ask a fundamental question: \emph{Can we automatically discover the visual structure within a diffusion model's knowledge of a concept?} Rather than requiring user-specified controls, we aim to decompose the model's internal representations into expressive directions that users can explore and combine.

To address these needs, we present \textbf{SliderSpace}, a framework that brings systematic explorability to diffusion models. Given just a text prompt, SliderSpace discovers a canonical set of meaningful, diverse, and controllable directions within the model's knowledge of that concept. Each direction is implemented as a low-rank adapter~\cite{hu2021lora} that can be scaled and composed with others, allowing users to explore and smoothly combine different aspects of variation, as shown in Figure~\ref{fig:intro}.

We ground SliderSpace discovery in three key requirements for meaningful decomposition of a diffusion model's visual manifold: 
\begin{enumerate}
    \item \textbf{Unsupervised Discovery:} The decomposition process should emerge from the intrinsic structure of the model's learned representation, rather than being guided by predefined attributes. This ensures we capture the true topology of the model's knowledge space rather than projecting our assumptions onto it.
    
    \item \textbf{Semantic Orthogonality:} Each discovered control must represent a distinct semantic direction. This is enforced in a semantic feature space, like CLIP, where every slider has an orthogonal effect in embeddings. This prevents discovering multiple controls that create similar semantic effects, making the system more efficient and easier.
    
    \item \textbf{Distribution Consistency:} Directions must induce consistent transformations across both random seeds and prompt variations. 
\end{enumerate}

These requirements naturally lead to our proposed framework, which we formalize in Section~\ref{sec:method}. As we show in our experiments, SliderSpace is architecture-agnostic, working with both conventional U-Net based models like Stable Diffusion~\cite{rombach2022high, rombach2022sd20, podell2023sdxl, turbo, dmd} and recent transformer-based architectures like Flux~\cite{flux}.

We demonstrate the expressiveness of SliderSpace through three applications: First, we show how SliderSpace can decompose high-level concepts into diverse and expressive components, revealing the natural axes of variation in the model's understanding. Second, we explore artistic style variation, where SliderSpace discovers directions that match or exceed the diversity of manually curated artist lists while being judged more useful by human evaluators. Finally, we show how SliderSpace can help reverse the mode collapse commonly observed in distilled diffusion models, restoring diversity while maintaining generation speed.

Beyond providing practical creative control, SliderSpace opens new avenues for understanding and utilizing the latent capabilities of diffusion models. By mapping these models' visual potential into intuitive, composable directions, we take a step toward making their creative possibilities more accessible and interpretable to users.

\section{Related Works}

Recent text-to-image diffusion models have demonstrated remarkable capabilities in generating diverse visual concepts \cite{rombach2022high}. While newer foundation models enhance text-image alignment through LLM-generated captions \cite{esser2024scaling, betker2023improving}, the fundamental challenge remains: text-conditioned generation is inherently under-determined, with multiple distinct outputs potentially satisfying the same prompt, making precise control challenging.

Prior work has explored various approaches to enhance generation control. One direction introduces additional conditioning modalities: spatial signals via adapters (ControlNet \cite{zhang2023addingconditionalcontroltexttoimage}), attention-based regional control \cite{chen2024trainingfreeregionalpromptingdiffusion}, and image-based conditioning for identity preservation \cite{ye2023ipadaptertextcompatibleimage, li2023photomaker, shi2024instantbooth} or style transfer via attention manipulation \cite{hertz2024stylealignedimagegeneration}. Another line of research focuses on refining existing images through disentangled text-driven editing \cite{brooks2023instructpix2pix, wu2024turboedit, deutch2024turboedittextbasedimageediting, xu2023infedit, brack2024ledits++, parmar2023zero}, where specific attributes are modified while preserving others, often using spatial masks. However, these text-driven approaches inherit the same under-determination challenges as the base models.

Notably, disentangled control has been more naturally achieved in GANs \cite{gans}, particularly through StyleGAN's low-dimensional latent space \cite{karras2019style}, which exhibits emergent disentanglement even at scale \cite{kang2023scaling}. This has enabled powerful latent space and image editing capabilities \cite{abdal2019image2stylegan, wu2021stylespace, abdal2021styleflow}. While diffusion models offer superior generation quality and diversity, they lack two key advantages of GANs' latent space: continuous, compositional edits and emergent disentanglement. This limitation prevents users from making serendipitous discoveries about visual variations captured in the training data, instead constraining them to variations they can explicitly describe through prompts.

Recent works have explored different approaches to discovering interpretable directions in diffusion models. The weights2weights method~\cite{dravid2024interpreting} learns a manifold of personalized model weights by fine-tuning individual LoRA adapters for each identity and applying PCA to discover a weight space that enables editing. While effective, this requires training separate models per instance. NoiseCLR~\cite{dalva2024noiseclr} learns text embeddings through contrastive learning on a data distribution, but the discovered directions can be arbitrary and lack semantic interpretability. Similarly, Liu et al. ~\cite{liu2023unsupervised} propose unsupervised decomposition of images into compositional concepts by learning multiple embeddings, but their approach often yields redundant or non-semantic directions. In contrast, our work directly learns a small set of semantically grounded sliders that decompose the distribution into interpretable and composable directions, enabling infinite creative variations through systematic exploration of the visual manifold.

\begin{figure*}
    \centering
    \includegraphics[width=\linewidth]{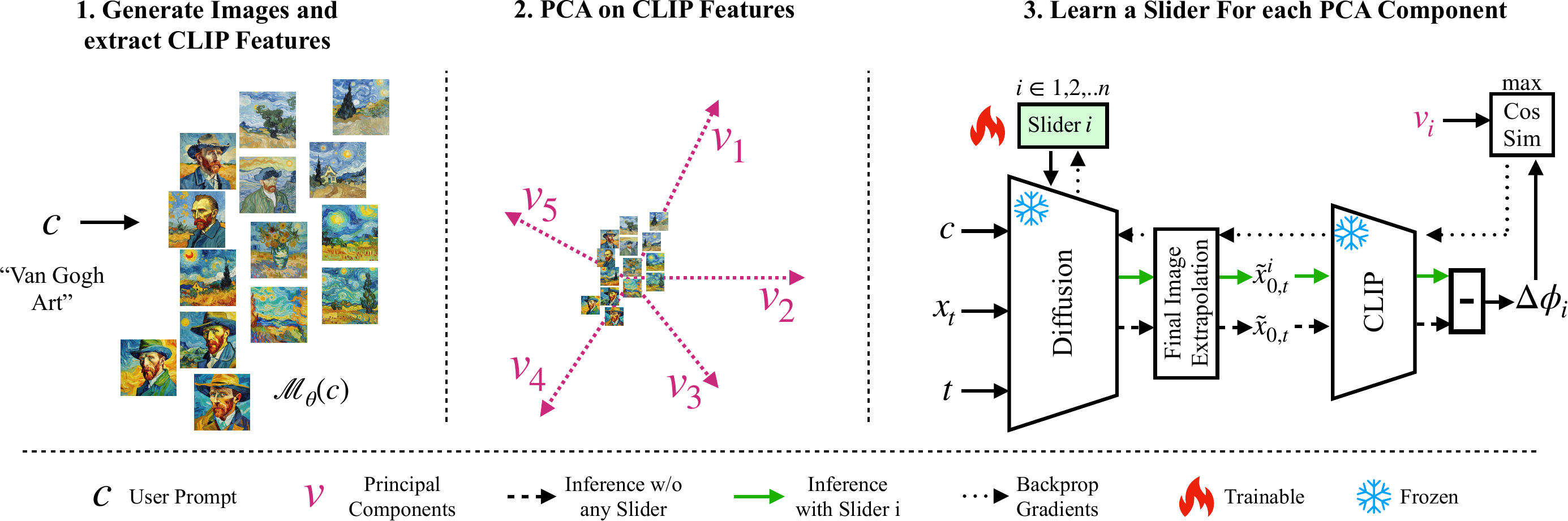}
    \caption{Given a prompt, SliderSpace first generates images and extracts the CLIP features. We then compute the spectral decompostion of the CLIP features and align the each slider with extracted principle components. Each slider is therefore trained to represent a unique semantic direction that are relevant in the diffusion model's knowledge of the prompt.}
    \label{fig:method}
\end{figure*}


Concept Sliders~\cite{gandikota2023concept} addressed continuous control by leveraging LoRA adaptors~\cite{hu2021lora} to learn user-defined attributes. Our work complements this by tackling emergent disentanglement through self-supervised decomposition of the model's inherent variations into composable control dimensions, enabling systematic exploration of the model's creative capabilities.

\section{Background}

\subsection{Latent Diffusion Models}

State-of-the-art text-to-image diffusion models~\cite{sd142022modelcard,sdv22022modelcard,podell2023sdxl, flux} often belong to the class of latent diffusion. Unlike traditional diffusion models that operate in pixel space, latent diffusion models work in a compressed latent space, offering significant computational advantages \cite{rombach2022high}. The diffusion modelling can be formalized as follows:

Let $\mathbf{x}_0$ be an initial image and $\mathbf{x}_T$ be pure Gaussian noise. The forward diffusion process gradually adds noise to the image. The generative process aims to reverse this diffusion, starting from $x_T$ and progressively denoising to reconstruct $\mathbf{x}_0$. At a timestep $t$ the model takes $\mathbf{x}_t$ as input and predicts a noise $\epsilon_t$ such that the next step $\mathbf{x}_{t-1}$:
\begin{equation}
\label{eq:xt}
\mathbf{x}_{t-1} \gets \frac{\mathbf{x}_t - \sqrt{1-\alpha_t}\epsilon_t}{\sqrt{\alpha_t}}
\end{equation}
We can estimate the final image $\Tilde{\mathbf{x}}_{0,t}$ by taking the same direction $\epsilon_t$ for remaining diffusion steps. This can be achieved by recursively applying the denoising from Equation~\ref{eq:xt} with the same direction. We label this ``Final Image Extrapolation'':
\begin{equation}
\label{eq:x0}
\Tilde{\mathbf{x}}_{0,t} \gets \frac{\mathbf{x}_t - \sqrt{1-\bar{\alpha_t}}\epsilon_t}{\sqrt{\bar{\alpha_t}}}.
\end{equation}
This enables us to visualize the final image $\Tilde{\mathbf{x}}_{0,t}$ that the diffusion model is planning of at each timestep $t$ without actually running denoising forward passes through all timesteps.

\subsection{LoRA: Low Rank Adaptors}
Low-rank adaptator (LoRA) \cite{hu2021lora} are a class of light-weight adaptors that are attachable to the weights of the model. Given a pre-trained model layer with weights $W_0 \in \mathbb{R}^{d \times k}$, LoRA decomposes the weight update $\Delta W$ as:

\begin{equation}
\Delta W = BA, \quad B \in \mathbb{R}^{d \times r}, A \in \mathbb{R}^{r \times k},
\end{equation}

\noindent where $r \ll \min(d,k)$ is a small rank that constrains the update to a low-dimensional subspace. This decomposition allows for efficient parameter updates and has shown success in various downstream tasks. However, the application of LoRA to unsupervised discovery of semantic directions in diffusion model weight space remains unexplored.
\section{Method}
\label{sec:method}
We present SliderSpace, a framework for decomposing a diffusion model's visual capabilities into semantically orthogonal control dimensions (Fig.~\ref{fig:intro}). Given a pre-trained text-to-image diffusion model $\theta$ and a prompt $c$, our goal is to discover $n$ independent directions that capture the principal modes of variation in the model's learned distribution.

\subsection{Problem Formulation}

Let $\mathcal{M}_\theta(c)$ denote the manifold of possible images that model $\theta$ can generate for prompt $c$. We aim to identify a set of controllable directions $\{\mathcal{T}_i\}_{i=1}^n$ that:
(1) span the major modes of variation in $\mathcal{M}_\theta(c)$,
(2) maintain semantic consistency across different initializations, and
(3) are mutually orthogonal in semantic space.

Building on recent advances in model adaptation~\cite{gandikota2023concept}, we formulate each control dimension as a LoRA adaptor~\cite{hu2021lora}, updating $\mathcal{T}_i$, where $i \in \{1, ..., n\}$. These lightweight adapters introduce targeted modifications to the model's cross-attention layers, enabling precise control over specific generative attributes. 
\subsection{SliderSpace: Unsupervised Visual Discovery}
SliderSpace discovery process consists of three key steps as depicted in Figure~\ref{fig:method}:

\paragraph{Distribution Sampling} 
First, we generate a diverse set of samples $\{x_j\}_{j=1}^m$ from $\mathcal{M}_\theta(c)$ by varying the random seed (for stability, $m\approx5000$). For each sample, we extract the estimated final image $\tilde{\mathbf{x}}_{0,t}$ at each timestep $t$ using Eq.~\ref{eq:x0}.

\paragraph{Semantic Decomposition}
We map each sample to a semantic embedding space, like CLIP, $\phi(\tilde{\mathbf{x}}_{0,t})$ and compute the principal components $V=\{v_i\}_{i=1}^n$ of the resulting distribution. These components represent orthogonal directions of maximal variation in semantic space:
\begin{equation}
    V = \text{PCA}(\{\phi(\tilde{x}_{0,t})\}_{j,t})
\end{equation}

\paragraph{Slider Training}
For each principal direction $v_i$, we train a corresponding adapter $\mathcal{T}_i$ to induce transformations that align with $v_i$ in semantic CLIP space. The training objective for each slider is:
\begin{equation}
\label{eq:contrast}
    \mathcal{L}_{\text{sliderspace}} = \sum_{i=1}^n 1 - \cos(\Delta\phi_i, v_i),
\end{equation}
where $\Delta\phi_i = \phi(\tilde{\mathbf{x}}_{0,t}^i) - \phi(\tilde{\mathbf{x}}_{0,t})$ represents the transformation induced by slider $i$ in embedding space and $\tilde{x}_{0,t}^i$ is the estimated final image with the slider $\mathcal{T}_i$ attached. This ensures that each slider's effect is semantically aligned with its corresponding principal direction while maintaining orthogonality with other sliders, building on established work showing that embedding differences encode semantic relationships~\cite{gal2022stylegan, patashnik2021styleclip}. 

Our formulation satisfies the three key requirements outlined in Section~\ref{sec:intro}. First, unsupervised discovery is achieved by deriving directions directly from the model's intrinsic variation through PCA in a semantic embedding space, without imposing predefined attributes or external supervision. Second, distribution consistency is enforced by our training objective $\mathcal{L}_{\text{sliderspace}}$, which ensures each slider's transformation maintains consistent direction in embedding space across different seeds and timesteps. Finally, semantic orthogonality is guaranteed through the PCA-based initialization of directions as each principal components are mutually orthogonal, ensuring each slider captures a distinct mode of variation. Together, these components enable the discovery of interpretable and reliable control dimensions that effectively decompose the model's learned distribution. In majority of this paper, we use CLIP~\cite{radford2021learning} as our primary semantic encoder.

\begin{figure*}[!ht]
    \centering
    \includegraphics[width=\linewidth]{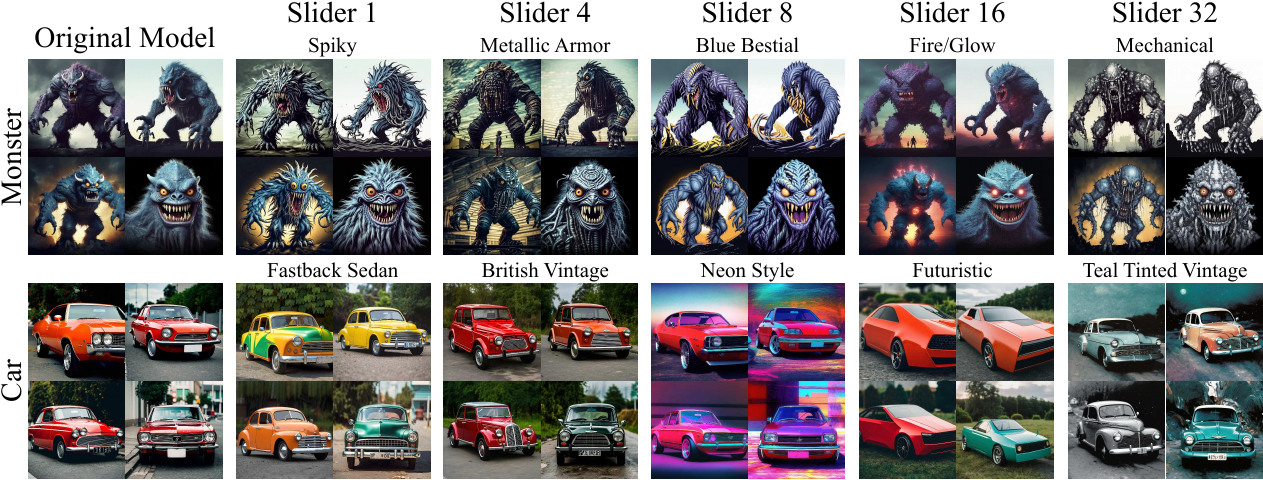}
    \caption{SliderSpace decomposes the visual variation of diffusion model's knowledge corresponding to a concept. These directions can be perceived as interpretable directions of the model's hierarchical knowledge. We show the decomposed slider direction for a concept using SliderSpace and the corresponding labels generated by Claude 3.5 Sonnet.}
    \label{fig:concept}
\end{figure*}

\subsection{Interpretability \& Control}

SliderSpace provides a dual contribution: it serves as both a framework for discovering expressive dimensions of control and as a mechanism for decomposing the model's learned concept space into semantically meaningful components. Each adapter functions as a ``slider'' controlling a specific attribute of the generated image, enabling fine-grained manipulation of the output while maintaining semantics.

The resulting set of adapters provides valuable insights into the model's conceptual understanding, revealing nuanced semantic relationships that may not be immediately apparent from the text prompt alone.

Moreover, the low-rank structure of our adapters ensures computational efficiency and minimal memory overhead, making them particularly suitable for real-time interactive applications. This enables users to explore the concept space dynamically by modulating the influence of each adapter, facilitating intuitive creative control over the image generation process while maintaining the underlying semantic integrity of the original prompt.
\section{Experiments}
We conduct our main experiments to evaluate SliderSpace using SDXL-DMD~\cite{dmd}, a 4-step distilled diffusion model. Our implementation requires less than 24GB VRAM and can discover 64 semantic directions in under 2 hrs on a single A100 GPU. When concepts exhibit severe mode collapse, the discovery process can be enhanced by generating data using undistilled base models or LLM-expanded prompts for increased sample diversity. We demonstrate SliderSpace's generalization to SDXL~\cite{podell2023sdxl}, SDXL-Turbo~\cite{turbo}, and transformer-based FLUX Schnell~\cite{flux} in appendix. Our analysis focuses on three key applications: concept decomposition, art styles exploration, and diversity enhancement in distilled models.


\subsection{Concept Decomposition} \label{sec:concept_exp}
We first demonstrate how SliderSpace can serve as an exploratory tool by decomposing high-level concepts into semantic directions that align with the diffusion model's internal representations. Summarizing and exposing the dominant variations a model is capable of for a particular prompt.

\begin{figure*}
    \centering
    \begin{tikzpicture}[baseline, xshift=-0.3\textwidth]
    \begin{axis}[
            width=0.45\textwidth,
            height=0.2\textwidth,
            ylabel={Diversity},
            ybar=2pt,   
            bar width=8pt,
            symbolic x coords={Mountain, Monster, Person, Van Gogh, Dog, Cat},
            xtick=data,
            enlarge x limits=0.2,
            legend image code/.code={%
                \draw[#1, draw=none] (0cm,0cm) rectangle (0.3cm,0.2cm);
            },
            legend style={
                at={(0.9,-0.05)},  
                anchor=north west,  
                draw=none,         
                fill=none,         
                legend columns=1
            },
            ymin=0,
            ymajorgrids=true,
            grid style=dashed,
            x tick label style={
                rotate=30,
                anchor=east
            },        ]
        \addplot[red,fill=red!30] coordinates {
            (Mountain,0.217)
            (Monster,0.266)
            (Person,0.423)
            (Van Gogh,0.349)
            (Dog,0.246)
            (Cat,0.246)        };

        \addplot[blue,fill=blue!30] coordinates {
            (Mountain,0.387)
            (Monster,0.472)
            (Person,0.613)
            (Van Gogh,0.557)
            (Dog,0.444)
            (Cat,0.387)        };
        
        \end{axis}
    \end{tikzpicture}
    \hspace{0.02\textwidth}
    \begin{tikzpicture}[baseline, xshift=0.3\textwidth]
    \begin{axis}[
            width=0.45\textwidth,
            height=0.2\textwidth,
            ylabel={Text Alignment},
            ybar=2pt,   
            bar width=8pt,
            symbolic x coords={Mountain, Monster, Person, Van Gogh, Dog, Cat},
            xtick=data,
            enlarge x limits=0.2,
            legend image code/.code={%
                \draw[#1, draw=none] (0cm,0cm) rectangle (0.3cm,0.2cm);
            },
            legend style={
                at={(0.9,-0.05)},  
                anchor=north west,  
                draw=none,         
                fill=none,         
                legend columns=1
            },
            ymin=0,
            ymax=40.0,  
            ymin=0,
            ymajorgrids=true,
            grid style=dashed,
            x tick label style={
                rotate=30,
                anchor=east
            },        ]
        \addplot[red,fill=red!30] coordinates {
            (Mountain,29.75)
            (Monster,28.14)
            (Person,27.38)
            (Van Gogh,31.94)
            (Dog,29.38)
            (Cat,29.95)        };

        \addplot[blue,fill=blue!30] coordinates {
            (Mountain,29.33)
            (Monster,28.14)
            (Person,27.40)
            (Van Gogh,32.20)
            (Dog,29.20)
            (Cat,29.20)        };
        
        \legend{SDXL-DMD, SliderSpace}
                
        \end{axis}
    \end{tikzpicture}    \caption{(left) SliderSpace generates diverse variations of a concept, as measured by DreamSim~\cite{dreamsim} distance across generated samples (higher is better). (right) The method maintains similar text-to-image alignment, as measured by CLIP Scores~\cite{hessel2021clipscore} (lower is better). }
    \label{fig:concept-diversity}
\end{figure*}
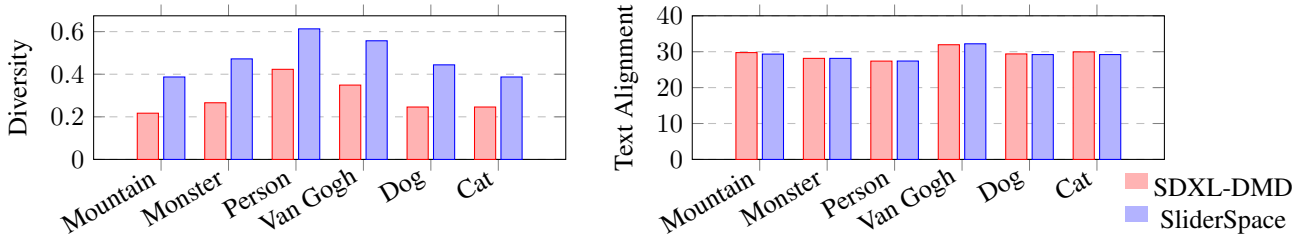

Given a concept prompt (e.g., "picture of a monster"), we discover SliderSpace directions.Figure~\ref{fig:concept} shows discovered sliders for ``Monster'', and ``car'' concepts. We label the sliders using Claude 3.5 Sonnet~\cite{claude} by showing multiple image pairs showing the effect of each slider and prompting to identify the semantic transformation being applied. Through these discovered directions, we demonstrate the manipulation of individual attributes. As these directions are intended to capture the model's visual possibilities for a prompt, we naturally ask the question, \textit{``How much variation is enabled by these directions?''}. To quantitatively evaluate the diversity enabled by SliderSpace against base model, we compute DreamSim \cite{dreamsim} distance to measure inter-image variation across 2500 generated samples per concept  (Figure~\ref{fig:concept-diversity}). For SliderSpace-augmented generation, we generate the images by randomly activating a sparse subset of 3 sliders (out of 32 discovered directions) for each generation. Our analysis reveals that SliderSpace-generated samples exhibit significantly higher inter-image diversity compared to the baseline model outputs. We measure the CLIP-Score~\cite{hessel2021clipscore} between the input prompts and the generated images to measure text-alignment with the prompt. We find that they have similar CLIP Scores to the images generated by the base model. This suggests that SliderSpace effectively expands the achievable variation in the model's knowledge, while maintaining semantic consistency. To validate our findings through human perception, we conducted pairwise comparisons of image grids generated by SliderSpace versus baseline methods. As shown in Table~\ref{tab:userconcept}, users consistently preferred SliderSpace outputs across diversity, utility, and creative potential.

\begin{table}
    \centering
    \small
    \begin{tabular}{rccc}
         & \multicolumn{3}{c}{\textbf{User Study (Win Rate \%)}} \\
        \textbf{Method vs.} & \textbf{``Diverse''} & \textbf{``Useful''}  & \textbf{``Creative''} \\
        \hline
        SDXL-DMD &  72.4 & 66.0 & 68.1 \\
        LLM + SDXL-DMD & 62.5 & 62.5 & 62.5 \\
        SDXL & 65.3 & 61.2 & 59.2  \\

    \end{tabular}
    \caption{Users perceive SliderSpace generated images to be more diverse, useful, and interesting. We show win-rate percentages of SliderSpace samples against baselines.}
    \label{tab:userconcept}
\end{table}

\subsection{Art Styles Exploration} \label{sec:art_exp}
We also evaluate to what degree SliderSpace can expose ``all'' of the art styles that the diffusion model has learned from its training data. As a proxy for ``all'', we use a diverse set of art styles, manually discovered and documented by ParrotZone~\cite{parrotzone}. We explore the visual artistic space by decomposing the prompt ``artwork in the style of a famous artist'' into 64 directions. This process enables us to discover the SliderSpace of art and create a comprehensive dictionary of discoverable art styles in a diffusion model.

\begin{figure*}
    \centering
    \includegraphics[width=\linewidth]{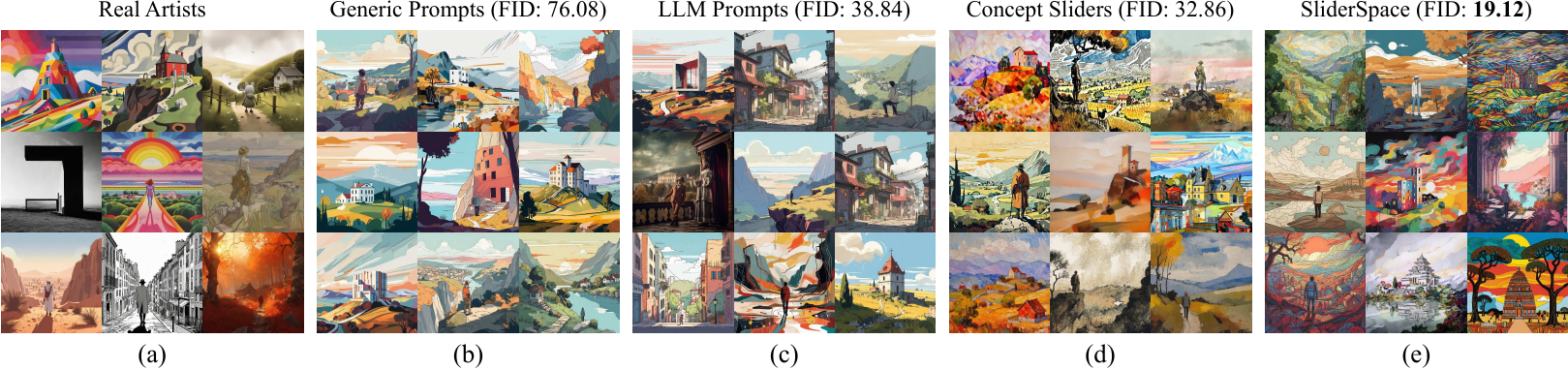}
    \caption{SliderSpace demonstrates broader artistic style coverage, as evidenced by the lower FID scores compared to both supervised Concept Sliders. Comparison of artistic style diversity using FID scores against reference distribution (a) derived from the complete artist dataset~\cite{parrotzone}. We compare against outputs from (b) generic art prompts (b), (c) LLM-generated art prompts, and (d) Concept Sliders.}
    \label{fig:artcompare}
\end{figure*}

\begin{figure*}[htbp]
    \centering
    \includegraphics[width=\linewidth]{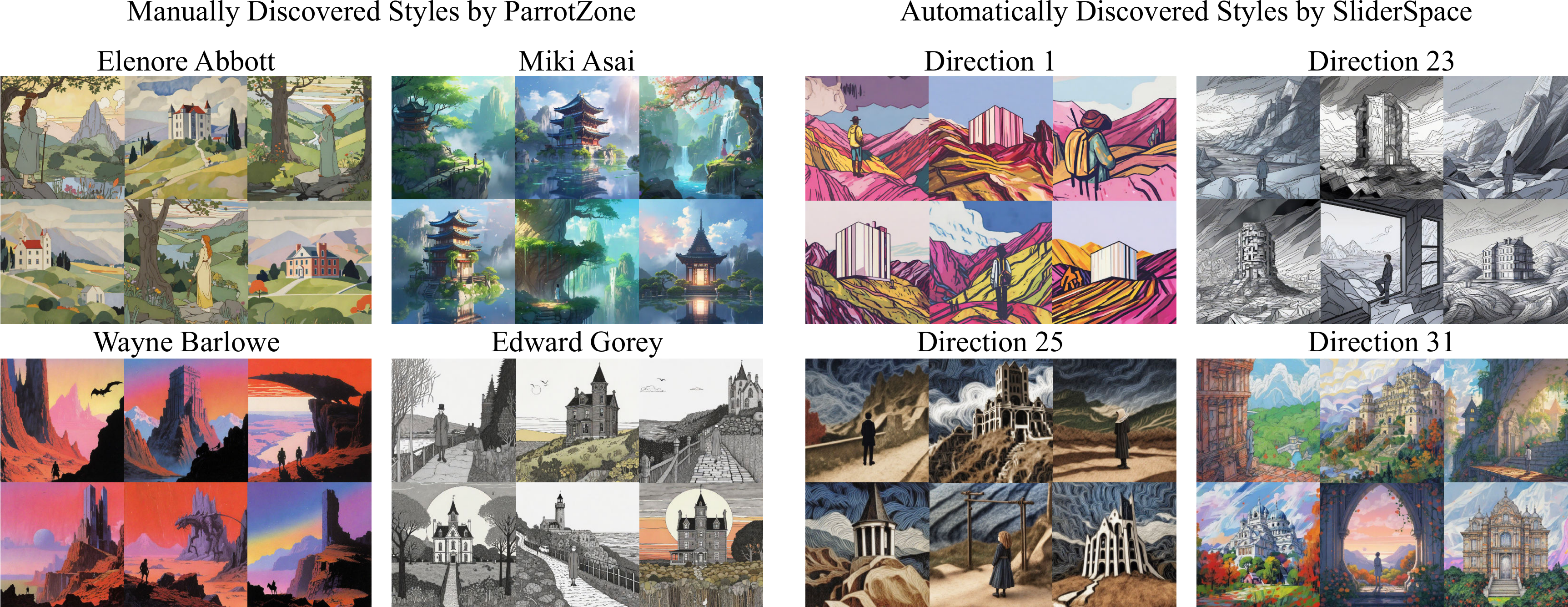}
    \caption{Comparison of artistic styles discovered in SDXL-DMD: (Left) Representative samples from artist-specific prompts manually curated by the Parrotzone community~\cite{parrotzone} through extensive exploration. (Right) Automatically discovered artistic directions using SliderSpace, which captures diverse and semantically meaningful variations without requiring explicit artist references.}
    \label{fig:artqualitative}
\end{figure*}

To compare how this measures up against supervised methods like Concept Slider~\cite{gandikota2023concept}, we train 64 manually curated concept sliders using LLM generated training prompts. We compute FID scores against the generated samples conditioned on actual artist names from ParrotZone~\cite{parrotzone} dataset. The dataset contains 4388 artists mimicked by SDXL, are discovered through exhaustive manual search. We also establish two other baselines. First, we use GPT/Claude to generate 4388 different artistic style names (e.g., ``cubism'', ``Warhol''). Second, we use a generic prompt like ``[subject] in the style of a famous artist''. Using our discovered SliderSpace, we randomly sample a sparse set of 3 sliders to generate an equivalent number of images. Specifically, for each baseline, we match the provided samples of \cite{parrotzone} by creating two images per style for ``building landscape'' and ``character portraits'' each (total of four images per style) to ensure same semantic structure across the datasets and that FID measures the artistic spread. Figure~\ref{fig:artcompare} shows random, non-cherry picked images from all the methods and their corresponding FID scores. We find that SliderSpace generates a distribution significantly more closely matches the manually curated artist names than the baselines, including supervised methods like Concept Sliders. Figure~\ref{fig:artqualitative} shows qualitative comparison of art styles discovered by SliderSpace and styles discovered manually ~\cite{parrotzone}.

To validate the practical utility of our ``artistic'' SliderSpace, we conduct a user study examining both usefulness and diversity of generated samples (Table~\ref{tab:userart}). We conduct comparitive study with 2 grids of 9 images (total of 1000 pairs), one with slider generated images and other with baselines. Users find the SliderSpace-generated images are far more diverse and useful than the baseline methods. We also find that users would prefer using SliderSpace images over images prompted with the real artist names~\cite{parrotzone}, while finding these both to be equally diverse. We provide more userstudy details and qualitative examples in Appendix.

\begin{table}
    \centering
    \small
    \begin{tabular}{rcc}
       & \multicolumn{2}{c}{\textbf{User study (win rate \%)}} \\
        \textbf{Method} & \textbf{``Diverse''} & \textbf{``Useful''}  \\
        \hline
        \,vs.\, Real Artist Prompts~\cite{parrotzone} &  54 & 63  \\
        \,vs.\, LLM Prompts & 73 & 67 \\
        \,vs.\, Generic Prompts & 87 & 88   \\

    \end{tabular}
    \caption{Pairwise comparison user study (win rates in \%) reveals that SliderSpace-extracted artistic styles achieve comparable diversity to manually curated artist~\cite{parrotzone}, while being significantly preferred for creative utilization. 
    }
    \label{tab:userart}
\end{table}

\subsection{Diversity Enhancement} \label{sec:diverse_exp}
\begin{figure*}
    \centering
    \includegraphics[width=.95\linewidth]{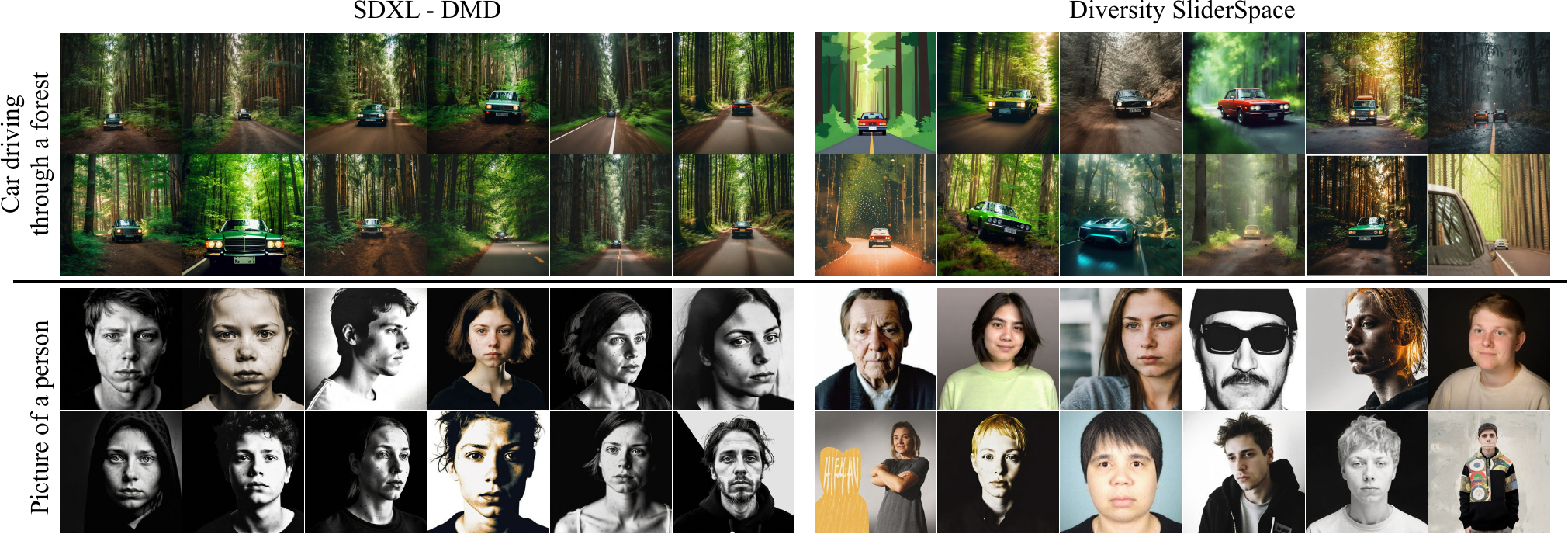}
    \caption{SliderSpace can also decompose general broad visual variation of diffusion model's and can be used to overcome model collapse in distilled models. We randomly sample a sparse set of sliders and generate the sample showing higher variation than base distilled model.}
    \label{fig:diverse}
\end{figure*}

Finally, we explore SliderSpace's ability to addressing mode collapse in distilled models. Instead of discovering SliderSpace for narrow distributions, we train it on a larger spread of 8000 randomly selected COCO-30k prompts and enhance them with LLM. We use DMD model for generating training dataset since we wish to undo mode collapse of DMD models. We discover a generic SliderSpace of 64 sliders that captures the generally applicable visual variations within SDXL-DMD. Figure ~\ref{fig:diverse} showcases qualitative examples of two distributions for the prompts ``Car driving through a forest'' and ``Picture of a person''. We demonstrate how generating images by randomly sampling from these SliderSpace directions effectively increases output diversity. This demonstrates that the mode collapse in distilled model can be reversed by discovering and exploring the visual structure. We show more examples in appendix.

In Table~\ref{tab:diverse}, we evaluate this improvement by measuring FID and CLIP scores on all COCO-30k prompts before and after applying SliderSpace to SDXL-DMD. We find that DMD-SliderSpace improves FID from the distilled version, almost matching the FID of undistilled SDXL. 
\begin{table}
    \centering
    \begin{tabular}{rcc}
        \textbf{Method} & \textbf{FID-30k ($\downarrow$)} & \textbf{CLIP ($\uparrow$)} \\
        \hline
        Real & - & 30.14 \\
        SDXL &  11.72 & 29.41 \\
        SDXL-DMD &  15.52 & 28.92 \\
        DMD-SliderSpace & 12.12 & 29.13 \\

    \end{tabular}
    \caption{SliderSpace trained on larger range of knowledge can improve the diversity of the distilled models (FID) while having a good text-image alignment (CLIP).}
    \label{tab:diverse}
\end{table}

\subsection{Slider Transferability}
\begin{figure*}[!ht]
    \centering
    \includegraphics[width=.95\linewidth]{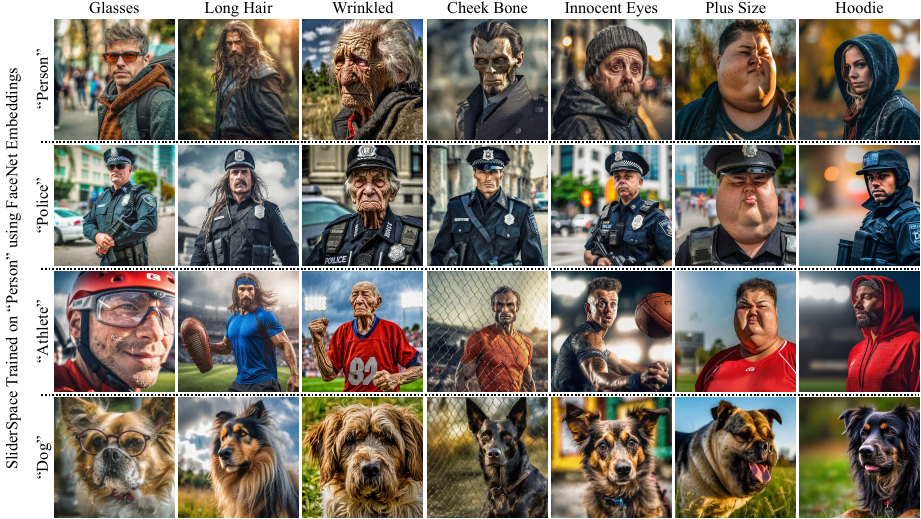}
    \caption{SliderSpace directions for the ``person'' concept successfully generalize to related ``police'' and ``athlete'' concepts. They also transfer to out-of-domain concepts like ``dog''}
    \label{fig:transfer}
\end{figure*}

Using FaceNet~\cite{schroff2015facenet} as the semantic encoder, we train SliderSpace on the concept ``person'' and discover interpretable directions controlling attributes like appearance and age (Figure~\ref{fig:transfer}). These directions not only transfer effectively to related concepts like ``police'' and ``athlete'' but also generalize surprisingly well to out-of-domain concepts like ``dog'', suggesting that SliderSpace captures fundamental visual transformations in the model's knowledge space.

Additional experiments analyzing hyperparameter choices (App.~\ref{sec:hyperparam}), alternative semantic embeddings (App.~\ref{sec:semantic}), and ablation studies (App.~\ref{sec:ablations}) are in Appendix.
\section{Limitations}
Our method's reliance on semantic embeddings introduces inherent biases present in encoder's training data. While these embeddings enable semantic consistency, they may not capture certain culturally-specific or nuanced artistic concepts. This highlights the need for more careful study on choices of the semantic embeddings and their effects on SliderSpace discovery. The current discovery process requires significant computational time ($\approx$ 2 hrs on A100), which may limit rapid experimentation and iteration. This computational overhead opens avenues for future research into training time optimizations. We also note that our method trains 4 times faster than  Concept Sliders for same number of sliders. For art style discovery, it is possible that the discovered directions are not one-to-one matched with the original artists. Further work can address discovery that nudges the directions to be aligned with real artists. 


\section{Conclusion}
SliderSpace is a simple framework that automatically decomposes diffusion models' capabilities into semantically meaningful and controllable directions. By leveraging spectral decomposition in semantic space combined with low-rank adaptation, our method enables systematic exploration of a model's latent creative space without requiring manual attribute specification. Through extensive experiments, we demonstrated SliderSpace's effectiveness across three key applications. First, our concept decomposition revealed interpretable variations within the model's knowledge representation, enabling fine-grained control while maintaining semantic consistency. Second, our exploration of artistic capabilities showed that SliderSpace can discover directions matching or exceeding the diversity of manually curated artist lists, while being rated more useful by human evaluators. Finally, we demonstrate how SliderSpace can help address mode collapse in distilled diffusion models, restoring diversity while preserving computational efficiency.

The ability of SliderSpace to uncover interpretable directions suggests that diffusion models may develop structured internal representations of visual concepts during training, without explicit supervision. By mapping these models' vast creative potential into intuitive, composable directions, our work takes a step toward making their capabilities more transparent and accessible.
\section*{Acknowledgment}
This work was done by RG at Adobe during their internship. RG and DB are supported by Open Philanthropy and NSF grant \#2403304.

\section*{Code}
Our methods are available as open-source code. Source code, trained sliderspace, and data sets for reproducing our results can be found at \href{https://sliderspace.baulab.info/}{\textcolor[rgb]{0.21,0.49,0.74}{sliderspace.baulab.info}} and at \href{https://github.com/rohitgandikota/sliderspace/}{\textcolor[rgb]{0.21,0.49,0.74}{github.com/rohitgandikota/sliderspace}}
.

\bibliographystyle{ieeenat_fullname}
\bibliography{main.bib}

\begin{thebibliography}{52}
\providecommand{\natexlab}[1]{#1}
\providecommand{\url}[1]{\texttt{#1}}
\expandafter\ifx\csname urlstyle\endcsname\relax
  \providecommand{\doi}[1]{doi: #1}\else
  \providecommand{\doi}{doi: \begingroup \urlstyle{rm}\Url}\fi

\bibitem[Abdal et~al.(2019)Abdal, Qin, and Wonka]{abdal2019image2stylegan}
Rameen Abdal, Yipeng Qin, and Peter Wonka.
\newblock Image2stylegan: How to embed images into the stylegan latent space?
\newblock In \emph{Proceedings of the IEEE/CVF international conference on computer vision}, pages 4432--4441, 2019.

\bibitem[Abdal et~al.(2021)Abdal, Zhu, Mitra, and Wonka]{abdal2021styleflow}
Rameen Abdal, Peihao Zhu, Niloy~J Mitra, and Peter Wonka.
\newblock Styleflow: Attribute-conditioned exploration of stylegan-generated images using conditional continuous normalizing flows.
\newblock \emph{ACM Transactions on Graphics (ToG)}, 40\penalty0 (3):\penalty0 1--21, 2021.

\bibitem[Anthropic(2024)]{claude}
Anthropic.
\newblock Introducing claude 3.5 sonnet, 2024.

\bibitem[BlackForestLabs(2024)]{flux}
BlackForestLabs.
\newblock Announcing state-of-the-art flux.1 dev and schnell models, 2024.

\bibitem[Brack et~al.(2024)Brack, Friedrich, Kornmeier, Tsaban, Schramowski, Kersting, and Passos]{brack2024ledits++}
Manuel Brack, Felix Friedrich, Katharia Kornmeier, Linoy Tsaban, Patrick Schramowski, Kristian Kersting, and Apolin{\'a}rio Passos.
\newblock Ledits++: Limitless image editing using text-to-image models.
\newblock In \emph{Proceedings of the IEEE/CVF Conference on Computer Vision and Pattern Recognition}, pages 8861--8870, 2024.

\bibitem[Brock et~al.(2018)Brock, Donahue, and Simonyan]{brock2018large}
Andrew Brock, Jeff Donahue, and Karen Simonyan.
\newblock Large scale gan training for high fidelity natural image synthesis.
\newblock \emph{arXiv e-prints}, pages arXiv--1809, 2018.

\bibitem[Brooks et~al.(2023)Brooks, Holynski, and Efros]{brooks2023instructpix2pix}
Tim Brooks, Aleksander Holynski, and Alexei~A Efros.
\newblock Instructpix2pix: Learning to follow image editing instructions.
\newblock In \emph{Proceedings of the IEEE/CVF Conference on Computer Vision and Pattern Recognition}, pages 18392--18402, 2023.

\bibitem[Chen et~al.(2024)Chen, Xu, Zheng, Dai, Wang, Zhang, Wang, and Zhang]{chen2024trainingfreeregionalpromptingdiffusion}
Anthony Chen, Jianjin Xu, Wenzhao Zheng, Gaole Dai, Yida Wang, Renrui Zhang, Haofan Wang, and Shanghang Zhang.
\newblock Training-free regional prompting for diffusion transformers.
\newblock \emph{arXiv preprint arXiv:2411.02395}, 2024.

\bibitem[Dalva and Yanardag(2024)]{dalva2024noiseclr}
Yusuf Dalva and Pinar Yanardag.
\newblock Noiseclr: A contrastive learning approach for unsupervised discovery of interpretable directions in diffusion models.
\newblock In \emph{Proceedings of the IEEE/CVF Conference on Computer Vision and Pattern Recognition}, pages 24209--24218, 2024.

\bibitem[Deutch et~al.(2024)Deutch, Gal, Garibi, Patashnik, and Cohen-Or]{deutch2024turboedittextbasedimageediting}
Gilad Deutch, Rinon Gal, Daniel Garibi, Or Patashnik, and Daniel Cohen-Or.
\newblock Turboedit: Text-based image editing using few-step diffusion models, 2024.

\bibitem[Dhariwal and Nichol(2021)]{dhariwal2021diffusion}
Prafulla Dhariwal and Alexander Nichol.
\newblock Diffusion models beat gans on image synthesis.
\newblock \emph{Advances in neural information processing systems}, 34:\penalty0 8780--8794, 2021.

\bibitem[Dravid et~al.(2024)Dravid, Gandelsman, Wang, Abdal, Wetzstein, Efros, and Aberman]{dravid2024interpreting}
Amil Dravid, Yossi Gandelsman, Kuan-Chieh Wang, Rameen Abdal, Gordon Wetzstein, Alexei~A Efros, and Kfir Aberman.
\newblock Interpreting the weight space of customized diffusion models.
\newblock \emph{arXiv preprint arXiv:2406.09413}, 2024.

\bibitem[Esser et~al.(2024)Esser, Kulal, Blattmann, Entezari, M{\"u}ller, Saini, Levi, Lorenz, Sauer, Boesel, et~al.]{esser2024scaling}
Patrick Esser, Sumith Kulal, Andreas Blattmann, Rahim Entezari, Jonas M{\"u}ller, Harry Saini, Yam Levi, Dominik Lorenz, Axel Sauer, Frederic Boesel, et~al.
\newblock Scaling rectified flow transformers for high-resolution image synthesis.
\newblock In \emph{Forty-first International Conference on Machine Learning}, 2024.

\bibitem[Fu et~al.(2024)Fu, Tamir, Sundaram, Chai, Zhang, Dekel, and Isola]{dreamsim}
Stephanie Fu, Netanel Tamir, Shobhita Sundaram, Lucy Chai, Richard Zhang, Tali Dekel, and Phillip Isola.
\newblock Dreamsim: Learning new dimensions of human visual similarity using synthetic data.
\newblock \emph{Advances in Neural Information Processing Systems}, 36, 2024.

\bibitem[Gal et~al.(2022)Gal, Patashnik, Maron, Bermano, Chechik, and Cohen-Or]{gal2022stylegan}
Rinon Gal, Or Patashnik, Haggai Maron, Amit~H Bermano, Gal Chechik, and Daniel Cohen-Or.
\newblock Stylegan-nada: Clip-guided domain adaptation of image generators.
\newblock \emph{ACM Transactions on Graphics (TOG)}, 41\penalty0 (4):\penalty0 1--13, 2022.

\bibitem[Gandikota et~al.(2024)Gandikota, Materzy{\'n}ska, Zhou, Torralba, and Bau]{gandikota2023concept}
Rohit Gandikota, Joanna Materzy{\'n}ska, Tingrui Zhou, Antonio Torralba, and David Bau.
\newblock Concept sliders: Lora adaptors for precise control in diffusion models.
\newblock In \emph{European Conference on Computer Vision}, pages 172--188. Springer, 2024.

\bibitem[Goodfellow et~al.(2014)Goodfellow, Pouget-Abadie, Mirza, Xu, Warde-Farley, Ozair, Courville, and Bengio]{gans}
Ian Goodfellow, Jean Pouget-Abadie, Mehdi Mirza, Bing Xu, David Warde-Farley, Sherjil Ozair, Aaron Courville, and Yoshua Bengio.
\newblock Generative adversarial nets.
\newblock \emph{Advances in neural information processing systems}, 27, 2014.

\bibitem[Grabe et~al.(2022)Grabe, Gonz{'a}lez-Duque, Risi, and Zhu]{grabe2022towards}
Imke Grabe, Miguel Gonz{'a}lez-Duque, Sebastian Risi, and Jichen Zhu.
\newblock Towards a framework for human-ai interaction patterns in co-creative gan applications.
\newblock In \emph{Joint Proceedings of the ACM IUI Workshops}, 2022.

\bibitem[H{\"a}rk{\"o}nen et~al.(2020)H{\"a}rk{\"o}nen, Hertzmann, Lehtinen, and Paris]{harkonen2020ganspace}
Erik H{\"a}rk{\"o}nen, Aaron Hertzmann, Jaakko Lehtinen, and Sylvain Paris.
\newblock Ganspace: Discovering interpretable gan controls.
\newblock \emph{Advances in neural information processing systems}, 33:\penalty0 9841--9850, 2020.

\bibitem[Hertz et~al.(2022)Hertz, Mokady, Tenenbaum, Aberman, Pritch, and Cohen-Or]{hertz2022prompt}
Amir Hertz, Ron Mokady, Jay Tenenbaum, Kfir Aberman, Yael Pritch, and Daniel Cohen-Or.
\newblock Prompt-to-prompt image editing with cross attention control.
\newblock \emph{arXiv preprint arXiv:2208.01626}, 2022.

\bibitem[Hertz et~al.(2024)Hertz, Voynov, Fruchter, and Cohen-Or]{hertz2024stylealignedimagegeneration}
Amir Hertz, Andrey Voynov, Shlomi Fruchter, and Daniel Cohen-Or.
\newblock Style aligned image generation via shared attention.
\newblock In \emph{Proceedings of the IEEE/CVF Conference on Computer Vision and Pattern Recognition}, pages 4775--4785, 2024.

\bibitem[Hessel et~al.(2021)Hessel, Holtzman, Forbes, Bras, and Choi]{hessel2021clipscore}
Jack Hessel, Ari Holtzman, Maxwell Forbes, Ronan~Le Bras, and Yejin Choi.
\newblock Clipscore: A reference-free evaluation metric for image captioning.
\newblock \emph{arXiv preprint arXiv:2104.08718}, 2021.

\bibitem[Hoffmann(2016)]{hoffmann2016modeling}
Oliver Hoffmann.
\newblock On modeling human-computer co-creativity.
\newblock In \emph{Knowledge, Information and Creativity Support Systems: Selected Papers from KICSS’2014-9th International Conference, held in Limassol, Cyprus, on November 6-8, 2014}, pages 37--48. Springer, 2016.

\bibitem[Hu et~al.(2021)Hu, Shen, Wallis, Allen-Zhu, Li, Wang, Wang, and Chen]{hu2021lora}
Edward~J Hu, Yelong Shen, Phillip Wallis, Zeyuan Allen-Zhu, Yuanzhi Li, Shean Wang, Lu Wang, and Weizhu Chen.
\newblock Lora: Low-rank adaptation of large language models.
\newblock \emph{arXiv preprint arXiv:2106.09685}, 2021.

\bibitem[I et~al.(2022)I, B, Erratica, and Young]{parrotzone}
Surea I, Proxima~Centauri B, Erratica, and Stephen Young.
\newblock Image synthesis style studies, 2022.

\bibitem[James~Betker(2023)]{betker2023improving}
et~al James~Betker.
\newblock Improving image generation with better captions.
\newblock \emph{OpenAI Reports}, 2023.

\bibitem[Kang et~al.(2023)Kang, Zhu, Zhang, Park, Shechtman, Paris, and Park]{kang2023scaling}
Minguk Kang, Jun-Yan Zhu, Richard Zhang, Jaesik Park, Eli Shechtman, Sylvain Paris, and Taesung Park.
\newblock Scaling up gans for text-to-image synthesis.
\newblock In \emph{Proceedings of the IEEE/CVF Conference on Computer Vision and Pattern Recognition}, pages 10124--10134, 2023.

\bibitem[Karras et~al.(2019)Karras, Laine, and Aila]{karras2019style}
Tero Karras, Samuli Laine, and Timo Aila.
\newblock A style-based generator architecture for generative adversarial networks.
\newblock In \emph{Proceedings of the IEEE/CVF conference on computer vision and pattern recognition}, pages 4401--4410, 2019.

\bibitem[Kumari et~al.(2023)Kumari, Zhang, Zhang, Shechtman, and Zhu]{kumari2022customdiffusion}
Nupur Kumari, Bingliang Zhang, Richard Zhang, Eli Shechtman, and Jun-Yan Zhu.
\newblock Multi-concept customization of text-to-image diffusion.
\newblock In \emph{Proceedings of the IEEE/CVF Conference on Computer Vision and Pattern Recognition}, pages 1931--1941, 2023.

\bibitem[Li et~al.(2024)Li, Cao, Wang, Qi, Cheng, and Shan]{li2023photomaker}
Zhen Li, Mingdeng Cao, Xintao Wang, Zhongang Qi, Ming-Ming Cheng, and Ying Shan.
\newblock Photomaker: Customizing realistic human photos via stacked id embedding.
\newblock In \emph{IEEE Conference on Computer Vision and Pattern Recognition (CVPR)}, 2024.

\bibitem[Liu et~al.(2023)Liu, Du, Li, Tenenbaum, and Torralba]{liu2023unsupervised}
Nan Liu, Yilun Du, Shuang Li, Joshua~B Tenenbaum, and Antonio Torralba.
\newblock Unsupervised compositional concepts discovery with text-to-image generative models.
\newblock In \emph{Proceedings of the IEEE/CVF International Conference on Computer Vision}, pages 2085--2095, 2023.

\bibitem[Parmar et~al.(2023)Parmar, Kumar~Singh, Zhang, Li, Lu, and Zhu]{parmar2023zero}
Gaurav Parmar, Krishna Kumar~Singh, Richard Zhang, Yijun Li, Jingwan Lu, and Jun-Yan Zhu.
\newblock Zero-shot image-to-image translation.
\newblock In \emph{ACM SIGGRAPH 2023 Conference Proceedings}, pages 1--11, 2023.

\bibitem[Patashnik et~al.(2021)Patashnik, Wu, Shechtman, Cohen-Or, and Lischinski]{patashnik2021styleclip}
Or Patashnik, Zongze Wu, Eli Shechtman, Daniel Cohen-Or, and Dani Lischinski.
\newblock Styleclip: Text-driven manipulation of stylegan imagery.
\newblock In \emph{Proceedings of the IEEE/CVF international conference on computer vision}, pages 2085--2094, 2021.

\bibitem[Podell et~al.(2023)Podell, English, Lacey, Blattmann, Dockhorn, M{\"u}ller, Penna, and Rombach]{podell2023sdxl}
Dustin Podell, Zion English, Kyle Lacey, Andreas Blattmann, Tim Dockhorn, Jonas M{\"u}ller, Joe Penna, and Robin Rombach.
\newblock Sdxl: Improving latent diffusion models for high-resolution image synthesis.
\newblock \emph{arXiv preprint arXiv:2307.01952}, 2023.

\bibitem[Radford(2015)]{radford2015unsupervised}
Alec Radford.
\newblock Unsupervised representation learning with deep convolutional generative adversarial networks.
\newblock \emph{arXiv preprint arXiv:1511.06434}, 2015.

\bibitem[Radford et~al.(2021)Radford, Kim, Hallacy, Ramesh, Goh, Agarwal, Sastry, Askell, Mishkin, Clark, et~al.]{radford2021learning}
Alec Radford, Jong~Wook Kim, Chris Hallacy, Aditya Ramesh, Gabriel Goh, Sandhini Agarwal, Girish Sastry, Amanda Askell, Pamela Mishkin, Jack Clark, et~al.
\newblock Learning transferable visual models from natural language supervision.
\newblock In \emph{International conference on machine learning}, pages 8748--8763. PMLR, 2021.

\bibitem[Rombach(2022)]{rombach2022sd20}
Robin Rombach.
\newblock Stable diffusion 2.0 release, 2022.

\bibitem[Rombach and Esser(2022{\natexlab{a}})]{sd142022modelcard}
Robin Rombach and Patrick Esser.
\newblock Stable diffusion v1-4 model card, 2022{\natexlab{a}}.

\bibitem[Rombach and Esser(2022{\natexlab{b}})]{sdv22022modelcard}
Robin Rombach and Patrick Esser.
\newblock Stable diffusion v2 model card, 2022{\natexlab{b}}.

\bibitem[Rombach et~al.(2022)Rombach, Blattmann, Lorenz, Esser, and Ommer]{rombach2022high}
Robin Rombach, Andreas Blattmann, Dominik Lorenz, Patrick Esser, and BjÃ¶rn Ommer.
\newblock High-resolution image synthesis with latent diffusion models.
\newblock In \emph{Proceedings of the IEEE Conference on Computer Vision and Pattern Recognition (CVPR)}, 2022.

\bibitem[Ruiz et~al.(2023)Ruiz, Li, Jampani, Pritch, Rubinstein, and Aberman]{ruiz2022dreambooth}
Nataniel Ruiz, Yuanzhen Li, Varun Jampani, Yael Pritch, Michael Rubinstein, and Kfir Aberman.
\newblock Dreambooth: Fine tuning text-to-image diffusion models for subject-driven generation.
\newblock In \emph{Proceedings of the IEEE/CVF conference on computer vision and pattern recognition}, pages 22500--22510, 2023.

\bibitem[Ryu(2023)]{Ryu_lora}
Simo Ryu.
\newblock Cloneofsimo/lora: Using low-rank adaptation to quickly fine-tune diffusion models.s.
\newblock \emph{GitHub}, 2023.

\bibitem[Sauer et~al.(2025)Sauer, Lorenz, Blattmann, and Rombach]{turbo}
Axel Sauer, Dominik Lorenz, Andreas Blattmann, and Robin Rombach.
\newblock Adversarial diffusion distillation.
\newblock In \emph{European Conference on Computer Vision}, pages 87--103. Springer, 2025.

\bibitem[Schroff et~al.(2015)Schroff, Kalenichenko, and Philbin]{schroff2015facenet}
Florian Schroff, Dmitry Kalenichenko, and James Philbin.
\newblock Facenet: A unified embedding for face recognition and clustering.
\newblock In \emph{Proceedings of the IEEE conference on computer vision and pattern recognition}, pages 815--823, 2015.

\bibitem[Shen et~al.(2020)Shen, Yang, Tang, and Zhou]{shen2020interfacegan}
Yujun Shen, Ceyuan Yang, Xiaoou Tang, and Bolei Zhou.
\newblock Interfacegan: Interpreting the disentangled face representation learned by gans.
\newblock \emph{IEEE transactions on pattern analysis and machine intelligence}, 44\penalty0 (4):\penalty0 2004--2018, 2020.

\bibitem[Shi et~al.(2024)Shi, Xiong, Lin, and Jung]{shi2024instantbooth}
Jing Shi, Wei Xiong, Zhe Lin, and Hyun~Joon Jung.
\newblock Instantbooth: Personalized text-to-image generation without test-time finetuning.
\newblock In \emph{Proceedings of the IEEE/CVF Conference on Computer Vision and Pattern Recognition}, pages 8543--8552, 2024.

\bibitem[Wu et~al.(2021)Wu, Lischinski, and Shechtman]{wu2021stylespace}
Zongze Wu, Dani Lischinski, and Eli Shechtman.
\newblock Stylespace analysis: Disentangled controls for stylegan image generation.
\newblock In \emph{Proceedings of the IEEE/CVF Conference on Computer Vision and Pattern Recognition}, pages 12863--12872, 2021.

\bibitem[Wu et~al.(2024)Wu, Kolkin, Brandt, Zhang, and Shechtman]{wu2024turboedit}
Zongze Wu, Nicholas Kolkin, Jonathan Brandt, Richard Zhang, and Eli Shechtman.
\newblock Turboedit: Instant text-based image editing.
\newblock \emph{ECCV}, 2024.

\bibitem[Xu et~al.(2024)Xu, Huang, Pan, Ma, and Chai]{xu2023infedit}
Sihan Xu, Yidong Huang, Jiayi Pan, Ziqiao Ma, and Joyce Chai.
\newblock Inversion-free image editing with natural language.
\newblock In \emph{Conference on Computer Vision and Pattern Recognition 2024}, 2024.

\bibitem[Ye et~al.(2023)Ye, Zhang, Liu, Han, and Yang]{ye2023ipadaptertextcompatibleimage}
Hu Ye, Jun Zhang, Sibo Liu, Xiao Han, and Wei Yang.
\newblock Ip-adapter: Text compatible image prompt adapter for text-to-image diffusion models.
\newblock \emph{arXiv preprint arXiv:2308.06721}, 2023.

\bibitem[Yin et~al.(2024)Yin, Gharbi, Park, Zhang, Shechtman, Durand, and Freeman]{dmd}
Tianwei Yin, Micha{\"e}l Gharbi, Taesung Park, Richard Zhang, Eli Shechtman, Fredo Durand, and William~T Freeman.
\newblock Improved distribution matching distillation for fast image synthesis.
\newblock \emph{arXiv preprint arXiv:2405.14867}, 2024.

\bibitem[Zhang et~al.(2023)Zhang, Rao, and Agrawala]{zhang2023addingconditionalcontroltexttoimage}
Lvmin Zhang, Anyi Rao, and Maneesh Agrawala.
\newblock Adding conditional control to text-to-image diffusion models.
\newblock In \emph{Proceedings of the IEEE/CVF International Conference on Computer Vision}, pages 3836--3847, 2023.

\end{thebibliography}

\clearpage
\appendix
\setcounter{page}{1}
\counterwithin{figure}{section}
\counterwithin{table}{section}
\counterwithin{equation}{section}

\maketitlesupplementary

\section{Principal Component Analysis}
\begin{figure}[!htpb]
    \centering
    \includegraphics[width=.8\linewidth]{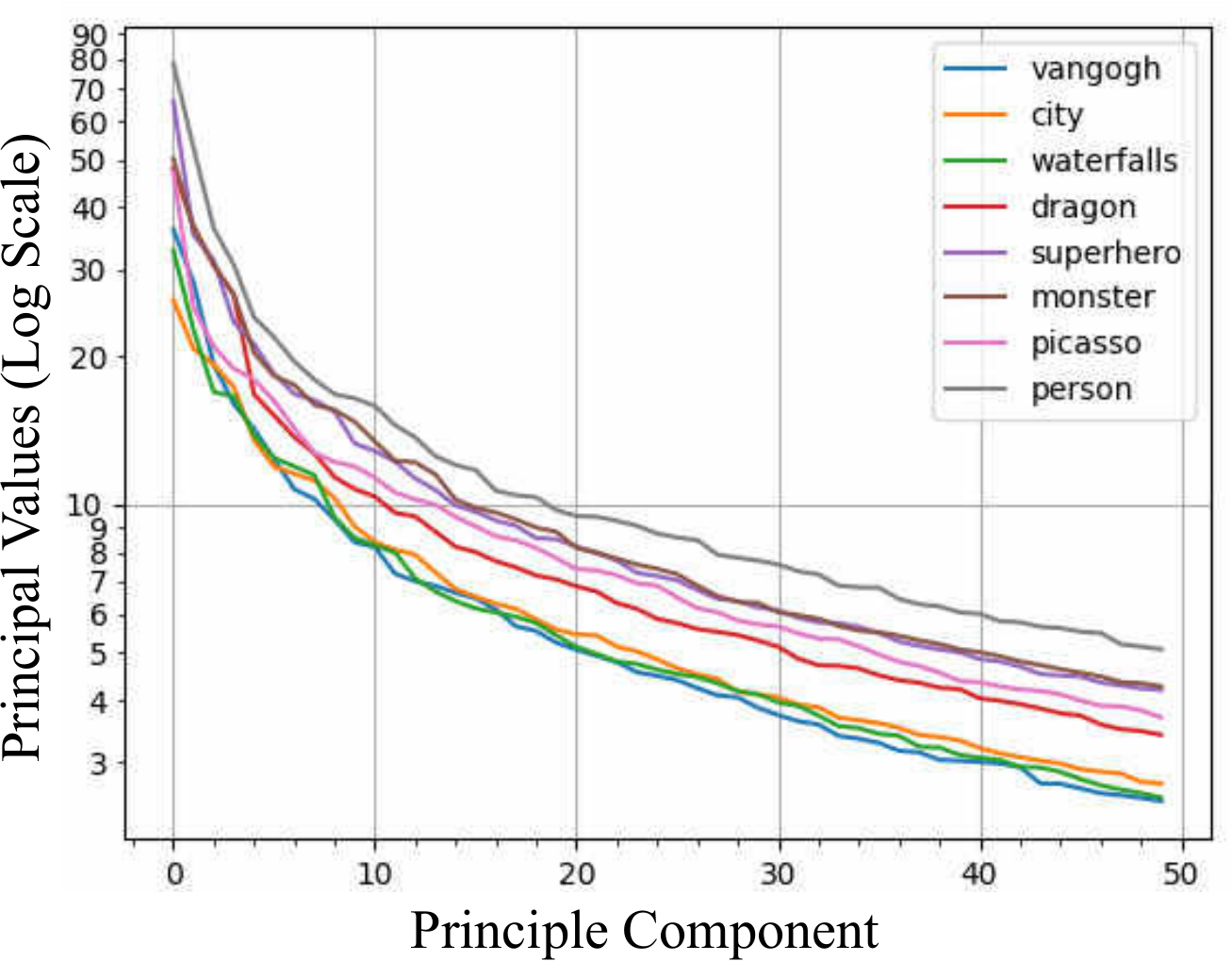}
    \caption{Common concepts like ``person'' show higher variation in CLIP space compared to rarer concepts like ``waterfalls''. 50th PCA component of ``person'' matches the 20th component of ``waterfalls,'' indicating the latter's more limited variation.} 
    \label{fig:pca}
\end{figure}

\begin{figure*}
    \centering
    \includegraphics[width=.8\linewidth]{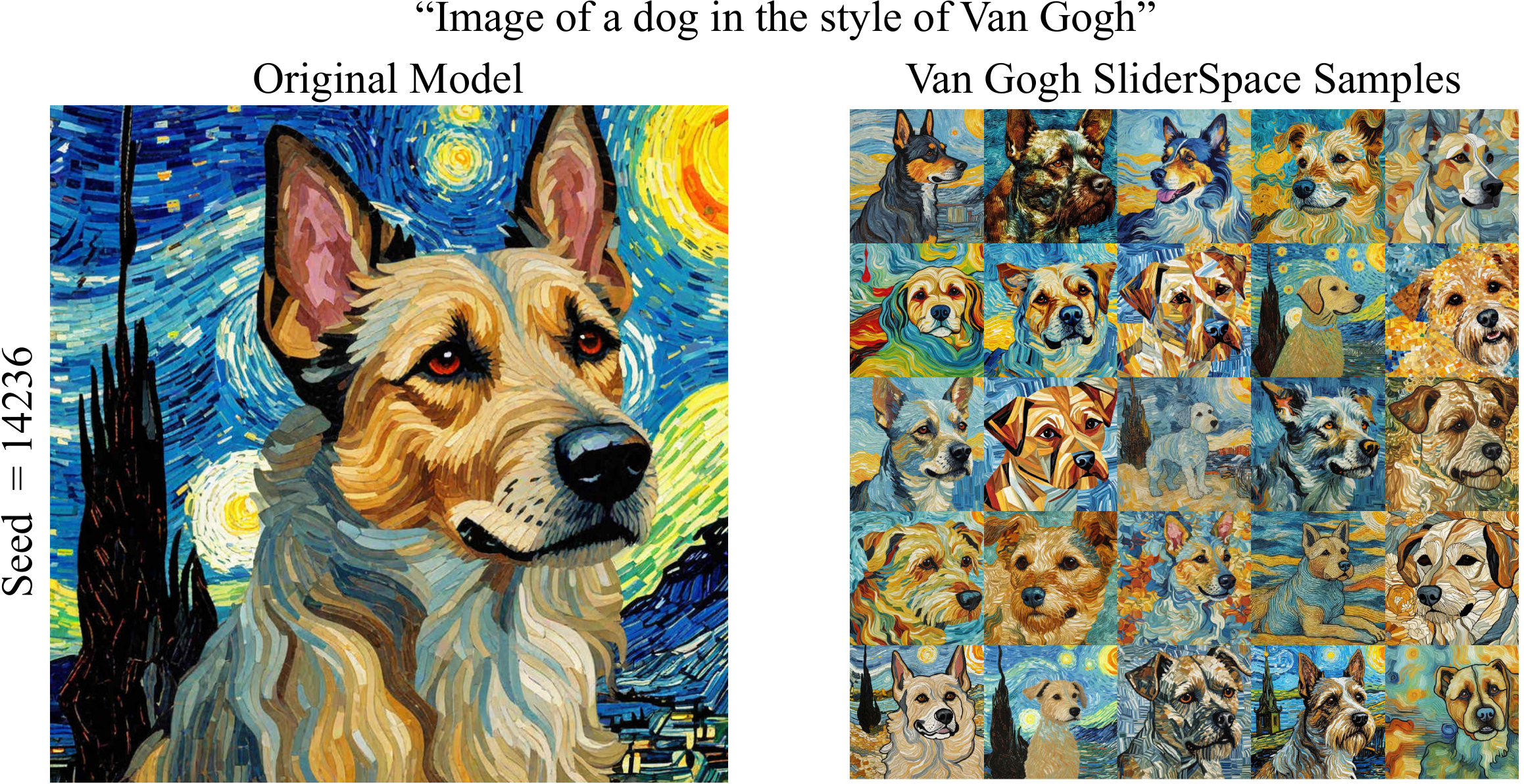}
    \caption{We show a few possible variations possible with SliderSpace directions. For a given seed and prompt, users can sample different combinations of sliders from SliderSpace and generate unique and diverse outputs (all variations from a single prompt and seed). We show this for the concept ``Van Gogh'' SliderSpace on SDXL-DMD2~\cite{dmd}.} 
    \label{fig:conceptdiverse1}
\end{figure*}

\begin{figure*}
    \centering
    \includegraphics[width=.8\linewidth]{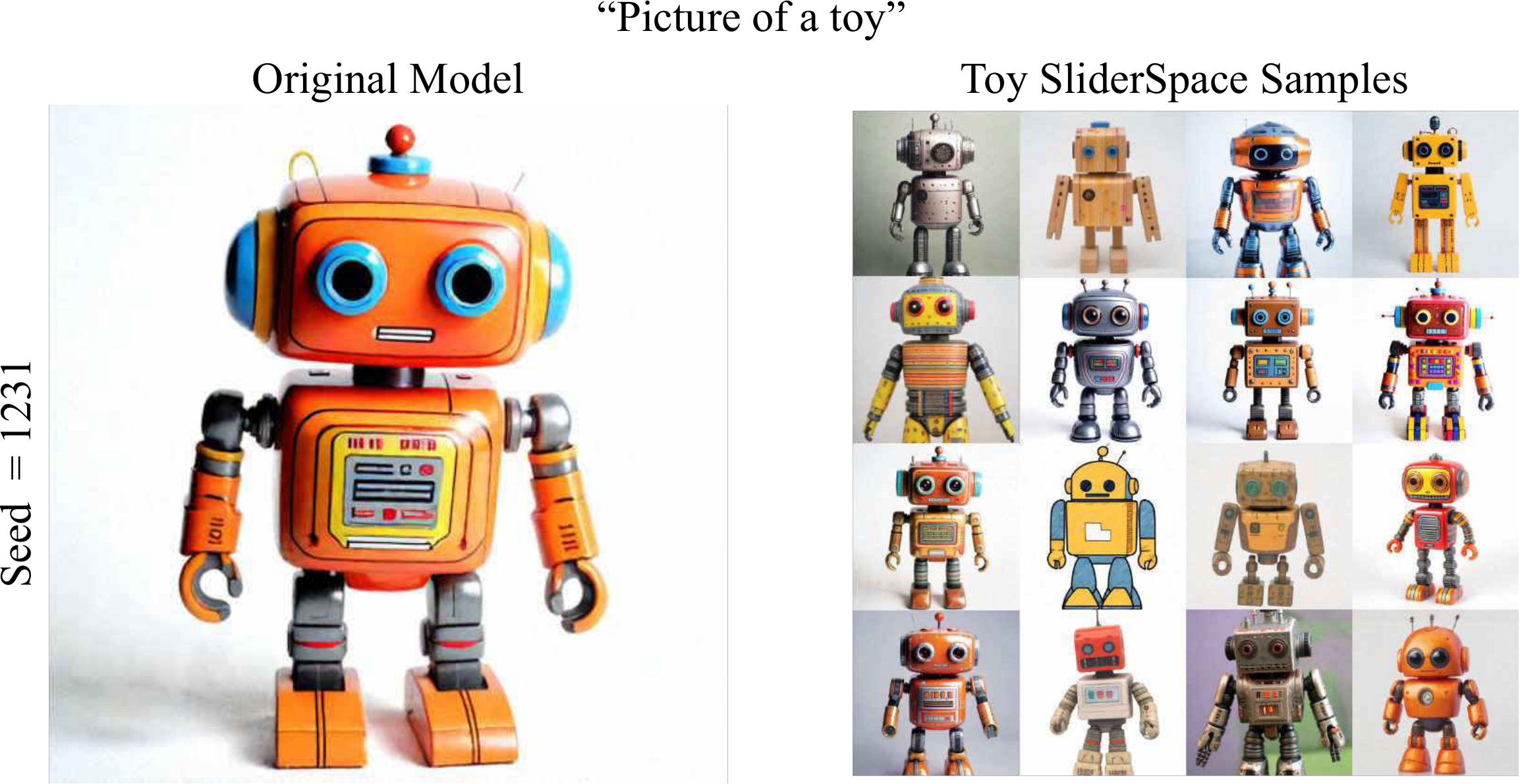}
    \caption{We show a few possible variations possible with SliderSpace directions. For a given seed and prompt, users can sample different combinations of sliders from SliderSpace and generate unique and diverse outputs (all variations from a single prompt and seed). We show this for the concept ``Toy'' SliderSpace on SDXL-DMD2~\cite{dmd}.} 
    \label{fig:conceptdiverse2}
\end{figure*}

Our analysis reveals that concepts frequently encountered in training data (e.g., "person") exhibit greater variation compared to concepts that are either less diverse or less common (e.g., "Van Gogh art" or "waterfalls"). We demonstrate this by analyzing the principal components for each concept through PCA visualization in Figure~\ref{fig:pca}. Notably, the 50th principal component for the "person" concept shows comparable variational magnitude to the 20th component of "waterfalls," highlighting the inherent variational differences across concepts. By discovering and uniformly sampling these variations, we effectively address the mode collapse problem in models, as shown in Figures~\ref{fig:conceptdiverse1} and~\ref{fig:conceptdiverse2}.
\vspace{-2em}

\section{Effect of TimeStep during Inference}
\begin{figure*}
    \centering
    \includegraphics[width=\linewidth]{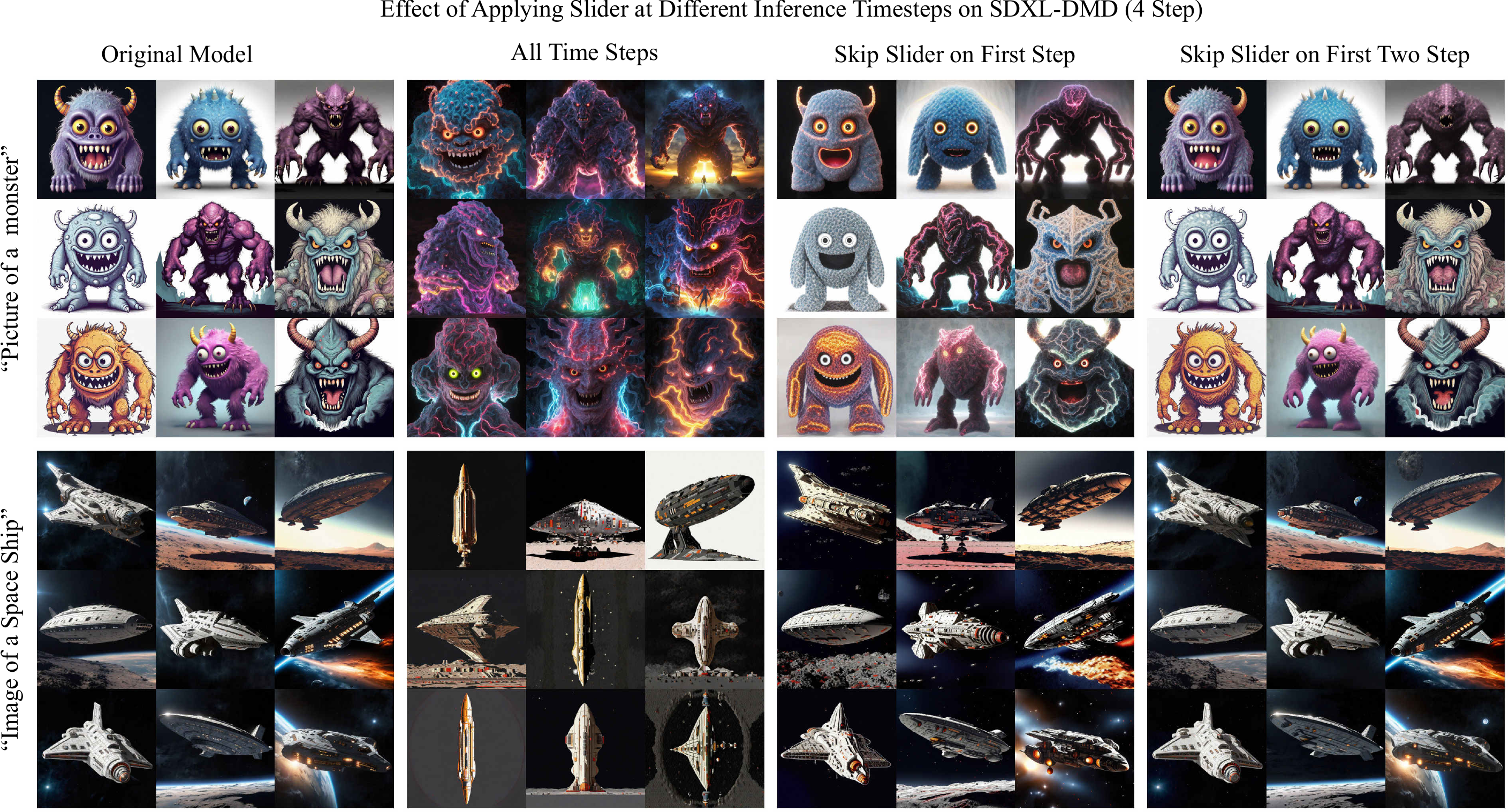}
    \caption{The choice of timestep at which sliders are applied can have an effect on the preciseness of the sliders. We show that when the sliders are applied to all the timesteps in inference, the images look different from the original models images for the same prompt and seed. But skipping the first timestep can lead to precise edits (similar observations as \citep{gandikota2023concept})} 
    \label{fig:timestep}
\end{figure*}

The temporal application of sliders during inference significantly impacts both the precision and magnitude of image edit(Fig.~\ref{fig:timestep}) for SDXL-DMD SliderSpace. When sliders are applied at all timesteps during inference, we observe strong semantic and structural changes in the generated image. But applying the slider after a few steps helps preserve the image structure while still enabling controlled edits. This latter approach facilitates more precise editing, albeit with subtler semantic alterations that can be amplified by increasing the slider strength parameter.


\subsection{Choice of Semantic Embeddings}
\label{sec:semantic}
\begin{figure*}[!ht]
    \centering
    \includegraphics[width=\linewidth]{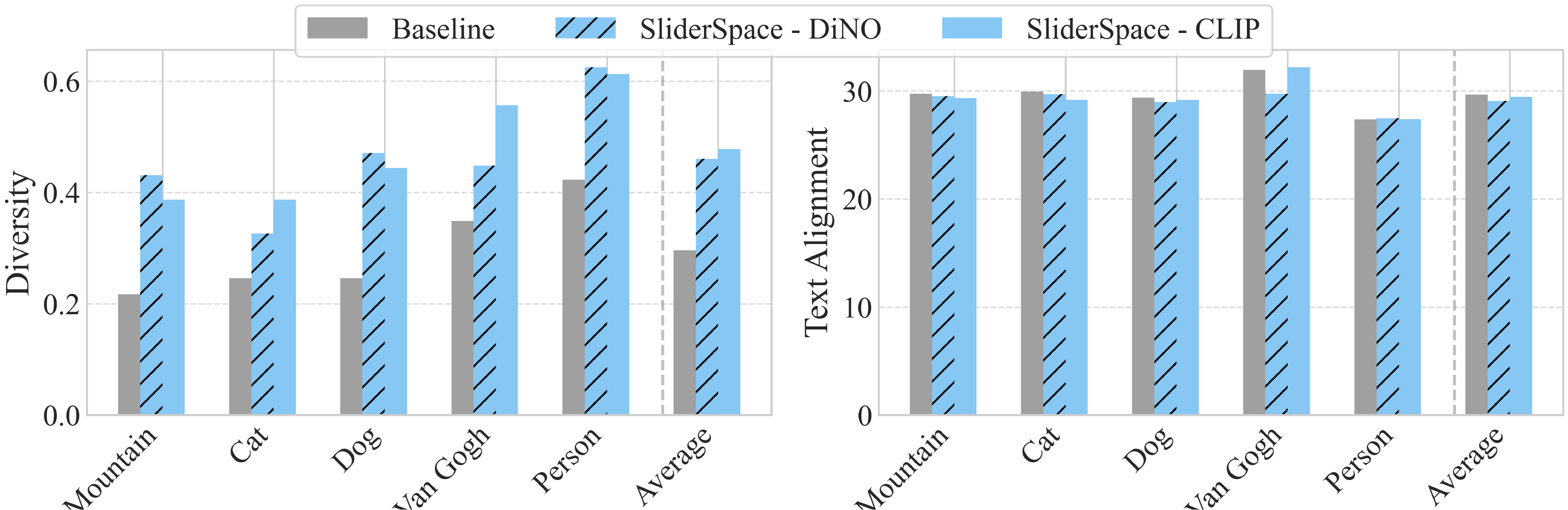}
    \vspace{-1.5em}
    \caption{SliderSpace shows similar diversity and text alignment when using either Dino-V2 or CLIP embeddings for PCA analysis.}
    \vspace{-0.5em}
    \label{fig:reb-concept}
\end{figure*}

While our primary implementation uses CLIP embeddings for semantic decomposition, SliderSpace is compatible with various semantic encoders. Our experiments with alternative embeddings like DINO-v2 and FaceNet demonstrate the framework's flexibility. As shown in Figure~\ref{fig:reb-concept}. DINO-v2 shows comparable overall performance to CLIP, with each encoder exhibiting different strengths across various concepts. For person-specific concepts, using FaceNet embeddings enables the discovery of fine-grained facial semantic directions as seen in Figure~\ref{fig:transfer}

The choice of encoder can be tailored to the target domain - CLIP for general concepts, DINO-v2 for certain visual attributes, and specialized encoders like FaceNet for domain-specific applications. This flexibility allows SliderSpace to adapt to different use cases while maintaining its core benefits of unsupervised discovery and semantic consistency.

\subsection{Hyperparameter Analysis}
\label{sec:hyperparam}
We analyze the impact of two key hyperparameters in SliderSpace: the number of PCA directions and the LoRA rank. Our experiments reveal that increasing PCA directions improves both knowledge coverage and output diversity up to about 40 dimensions, after which returns diminish. With just 10 directions, SliderSpace matches the FID scores of 64 manually created Concept Sliders when evaluated against artistic style distributions. Regarding model architecture, we find that lower-rank adaptors (particularly rank-one) efficiently capture variations with a fixed training budget, outperforming higher-rank versions while maintaining better FID scores than Concept Sliders across different ranks.

This analysis guides our choice of using rank-one adapters with 40 PCA directions as the default configuration, offering an optimal balance between performance and computational efficiency.

\begin{figure*}[!ht]
    \centering
    \includegraphics[width=\linewidth]{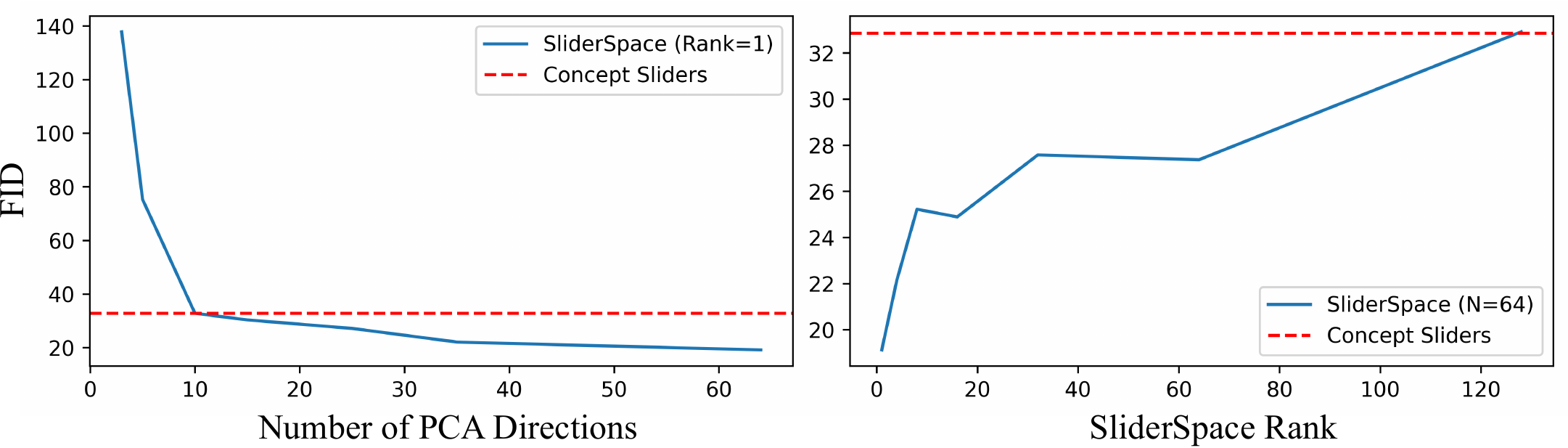}
    \vspace{-1.5em}
    \caption{Concept Sliders Comparison \& Hyperparameter analysis: (Left) Impact of PCA directions: SliderSpace with 10 directions matches the FID of 64 Concept Sliders. More directions, upto 40, leads to improved FID. (Right) Effect of LoRA rank: Given a fixed training budget rank-one sliders are efficient than higher rank versions and outperforms Concept Sliders}
    \vspace{-0.3em}
    \label{fig:reb-hyperparam}
\end{figure*}

\section{User Study}
\begin{figure*}
    \centering
    \includegraphics[width=\linewidth]{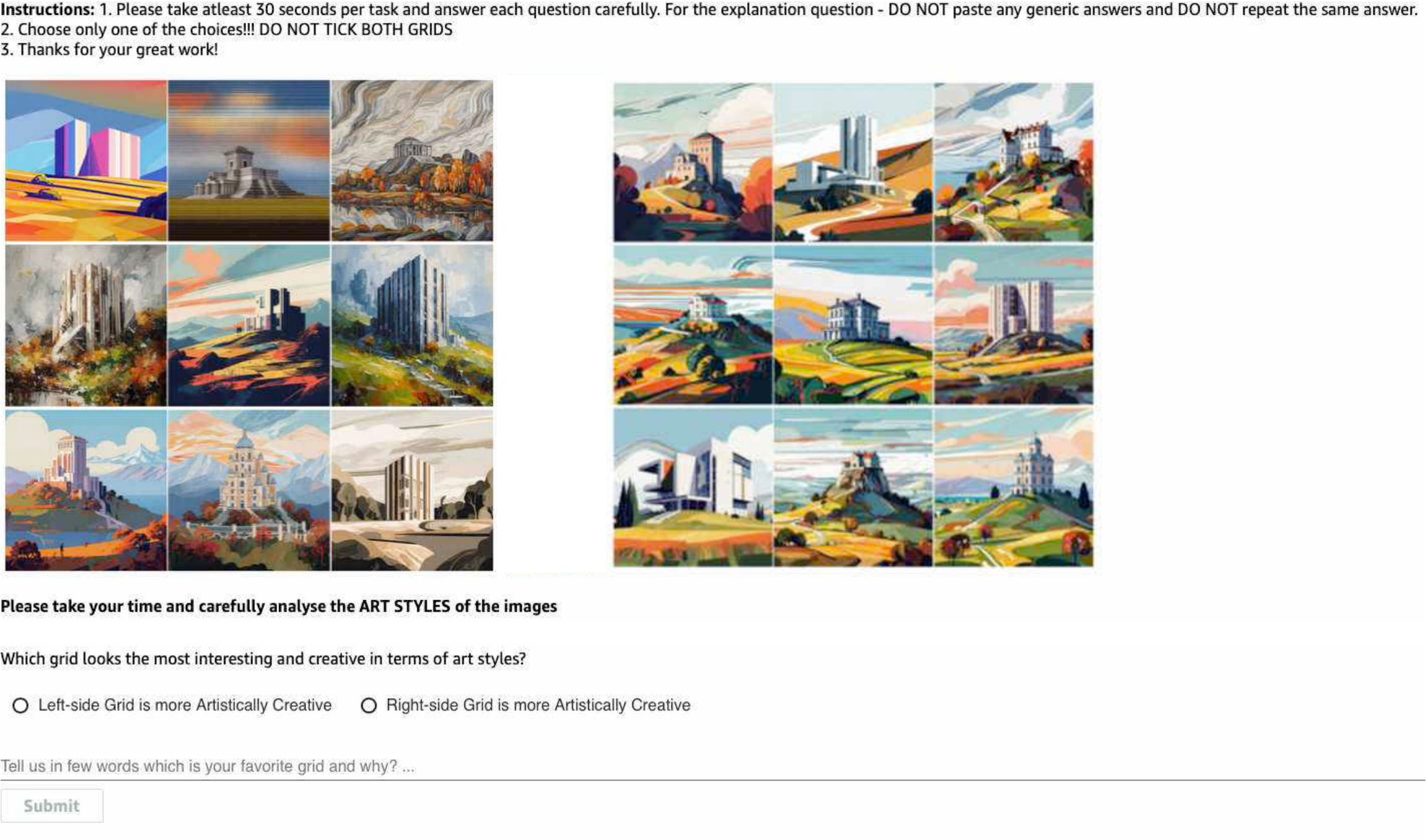}
    \caption{User study interface on Amazon Mechanical Turk. Users are shown images randomly sampled from SliderSpace or our baselines (Sec:~\ref{sec:art_exp}, and asked to identify the grid with most creative art renditions. } 
    \label{fig:user_art}
\end{figure*}

\begin{figure*}
    \centering
    \includegraphics[width=\linewidth]{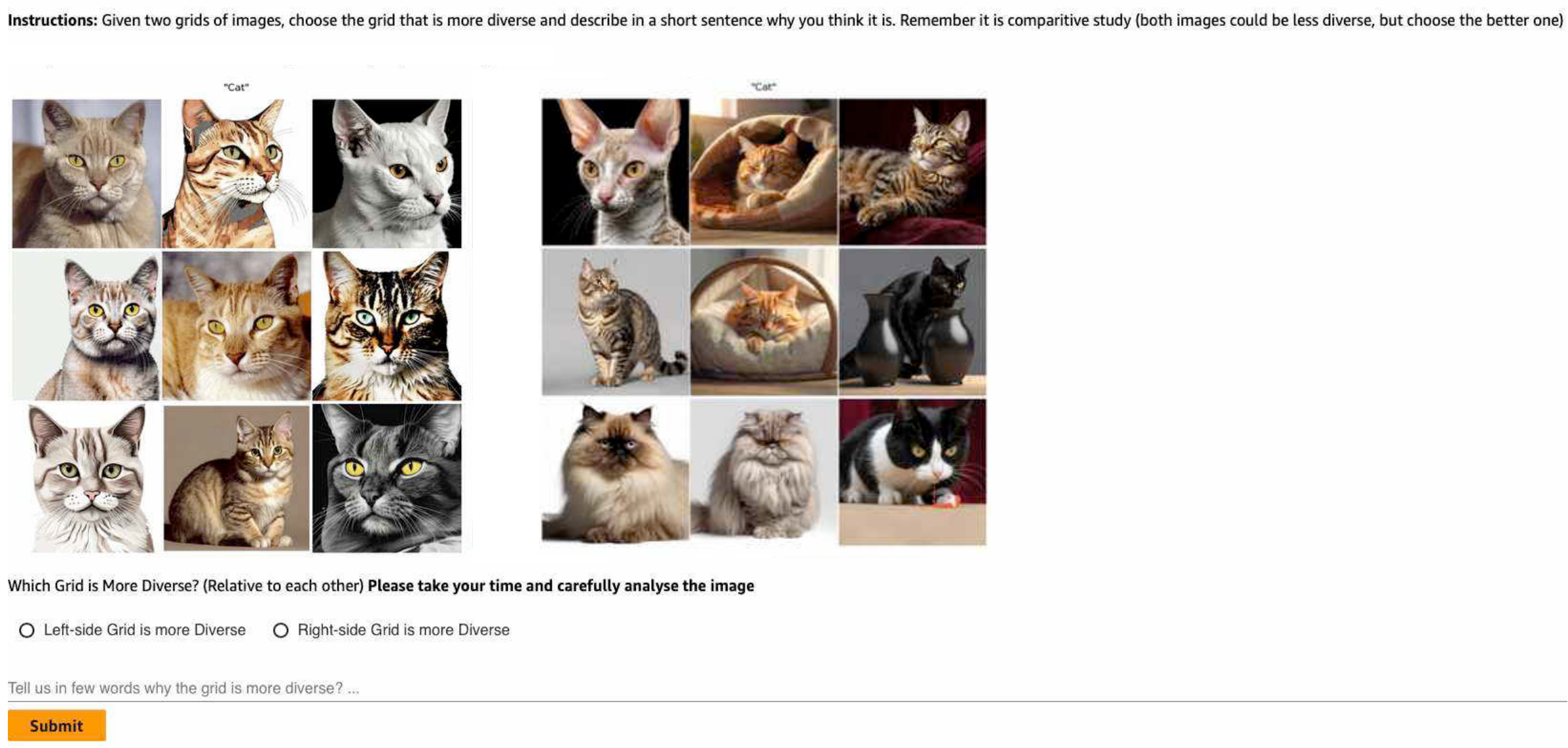}
    \caption{User study interface on Amazon Mechanical Turk. Users are shown images randomly sampled from SliderSpace or our baselines (Sec:~\ref{sec:concept_exp}, and asked to identify the grid with most diverse outputs. } 
    \label{fig:user_concept}
\end{figure*}

We conducted user studies to evaluate SliderSpace's effectiveness through Amazon MTurk. For artistic evaluation (Sec~\ref{sec:art_exp}), participants compared two 9-image grids - one generated by SliderSpace using 3 random sliders per image, and another by our baselines. Both sets used identical base prompts: ``a building in a stunning landscape'' and ``a character in a scenic environment''. As shown in Fig~\ref{fig:user_art}, participants rated which grid exhibited greater artistic diversity and utility for art applications. For conceptual evaluation (Sec~\ref{sec:concept_exp}), participants compared image grids based on diversity, generative utility, and creativity (Fig~\ref{fig:user_concept}). Grid presentation order was randomized across all experiments.

\section{Qualitative Results}
\subsection{Art Exploration}
\begin{figure*}[!htbp]
    \centering
    \includegraphics[width=.8\linewidth]{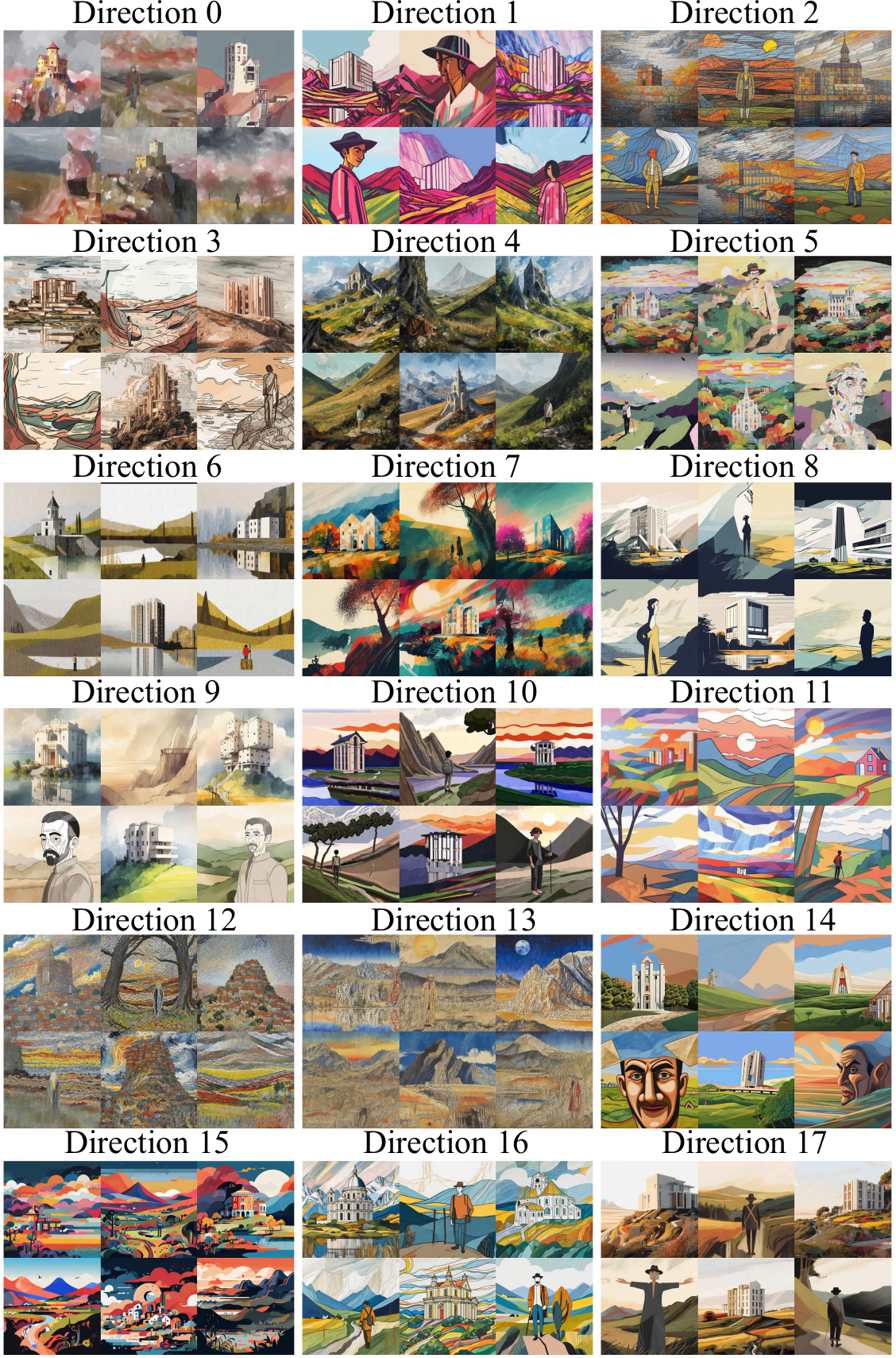}
    \caption{We show the top 18 art directions that are discovered in the SDXL-DMD2~\cite{dmd} SliderSpace for the concept ``art''.} 
    \label{fig:artdmd1}
\end{figure*}

\begin{figure*}[!htbp]
    \centering
    \includegraphics[width=.8\linewidth]{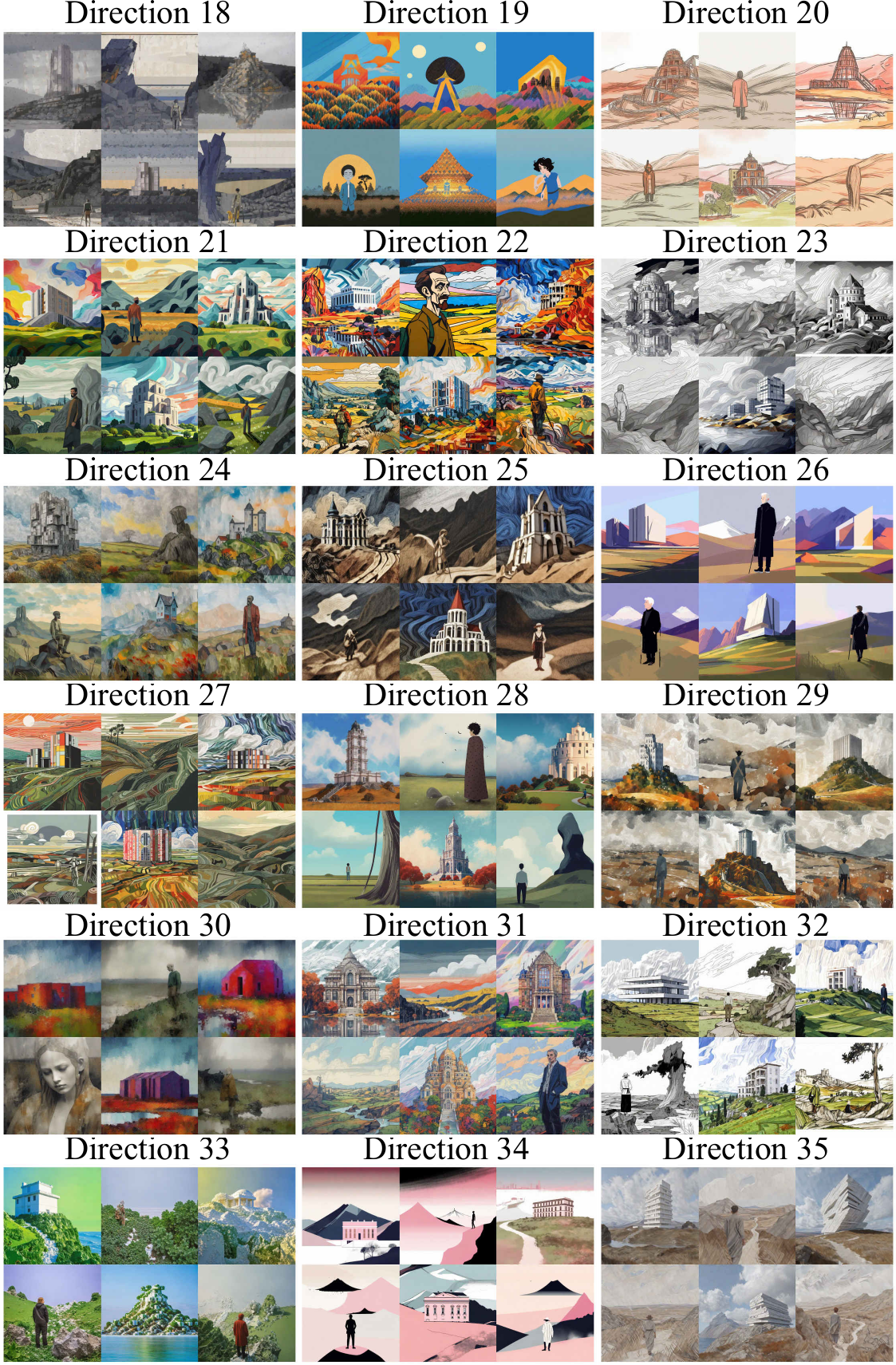}
    \caption{We show the top 18-36 art directions that are discovered in the SDXL-DMD2~\cite{dmd} SliderSpace for the concept ``art''.} 
    \label{fig:artdmd2}
\end{figure*}

\begin{figure*}[!htbp]
    \centering
    \includegraphics[width=.8\linewidth]{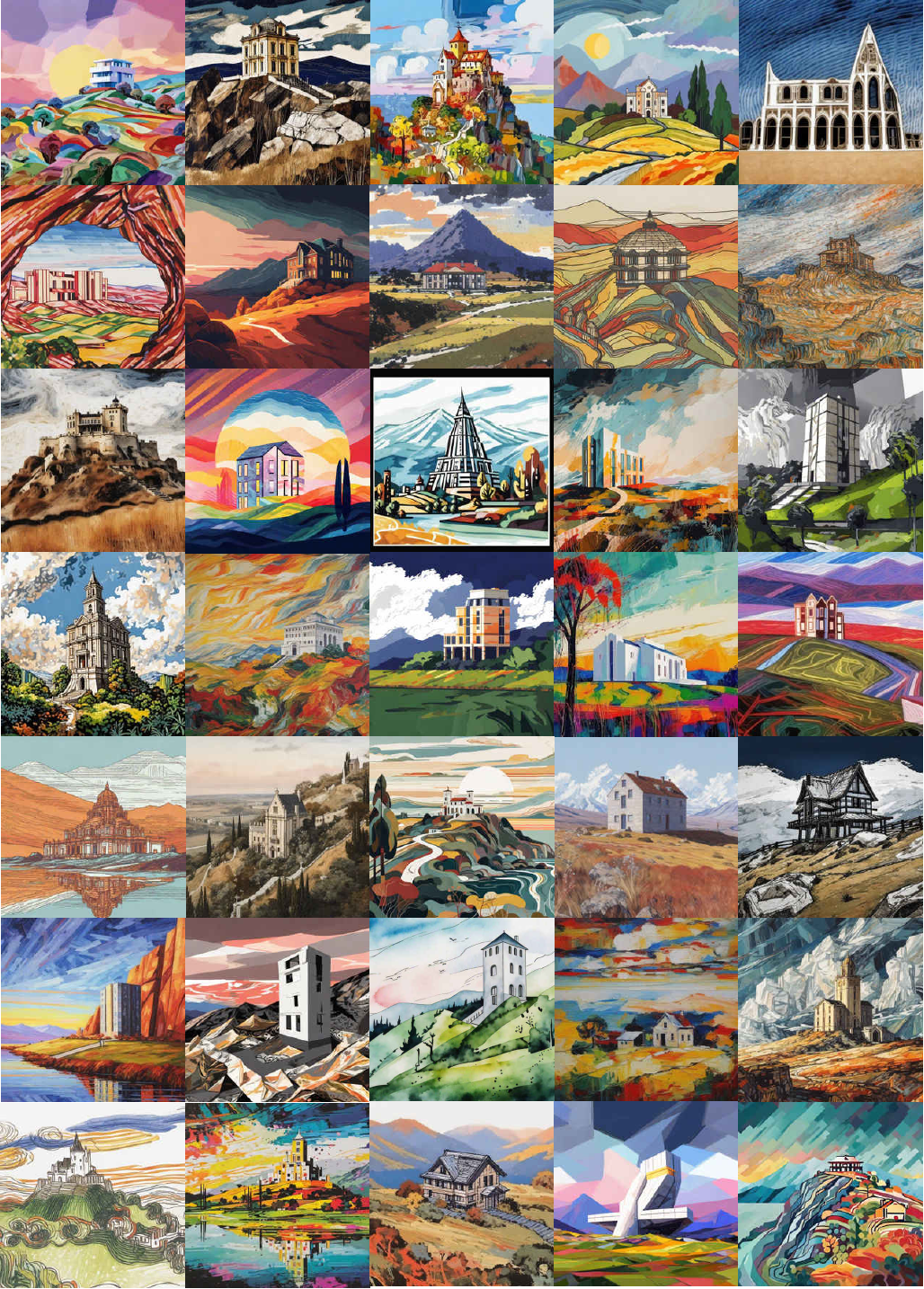}
    \caption{We show samples from our art experiments ~\ref{sec:art_exp}. We sample random 3 sliders from the SDXL-DMD2~\cite{dmd} SliderSpace for the concept ``art'' and generate images for the prompt ``a building in a stunning landscape the style of a famous artist''.} 
    \label{fig:artbuilding}
\end{figure*}

\begin{figure*}[!htbp]
    \centering
    \includegraphics[width=.8\linewidth]{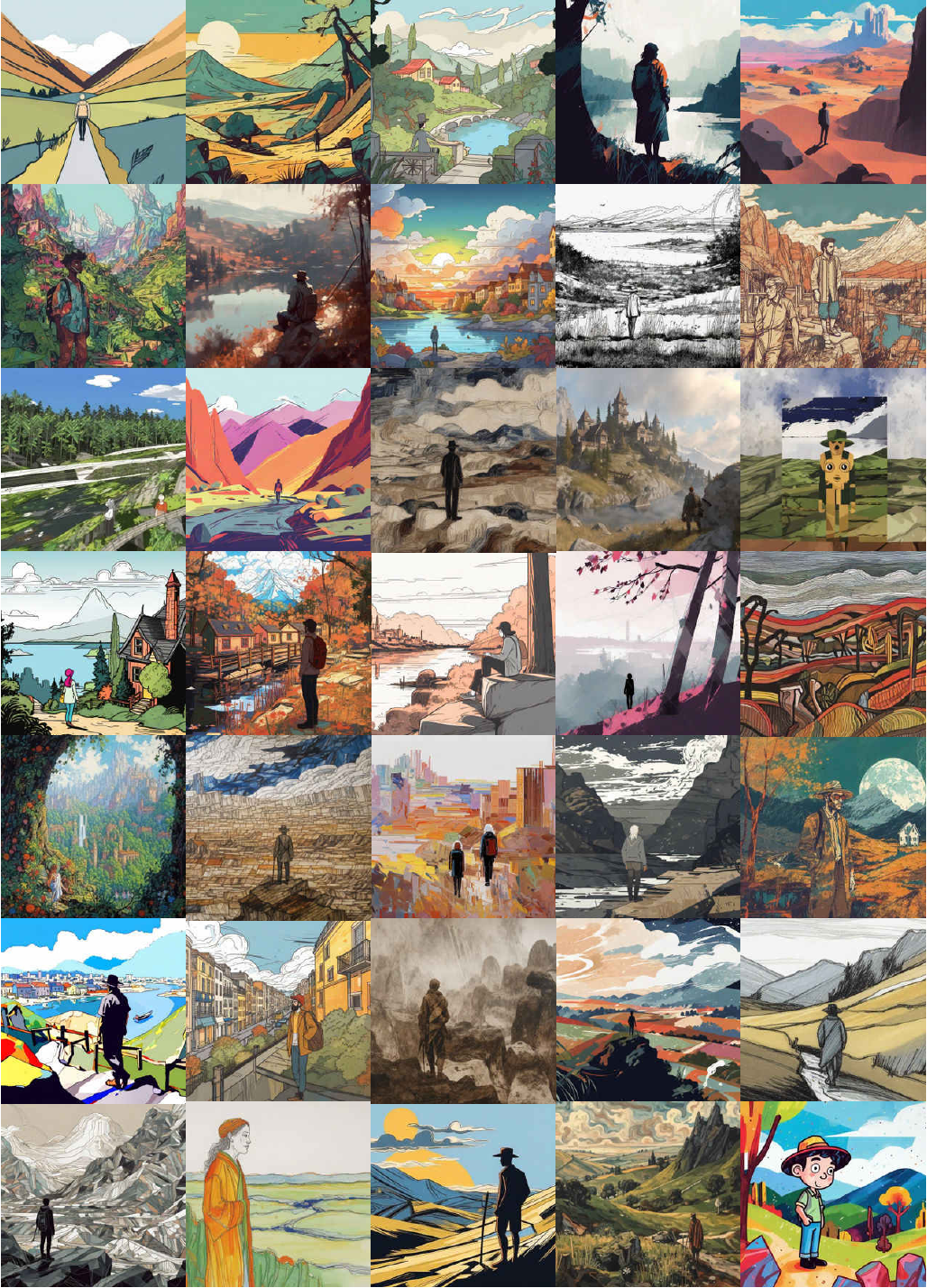}
    \caption{We show samples from our art experiments ~\ref{sec:art_exp}. We sample random 3 sliders from the SDXL-DMD2~\cite{dmd} SliderSpace for the concept ``art'' and generate images for the prompt ``a character in a scenic environment the style of a famous artist''.} 
    \label{fig:artcharacter}
\end{figure*}

We identify the top-36 distinct art directions discovered by SDXL-DMD2 SliderSpace for the concept "artwork in the style of famous artist" in Figures~\ref{fig:artdmd1} and~\ref{fig:artdmd2}. Additionally, we showcase various combinations of SliderSpace samples in Figures~\ref{fig:artbuilding} and~\ref{fig:artcharacter}, where we randomly sample three sliders to generate images for both characters and buildings (used in our art experiments and user studies in Section~\ref{sec:art_exp}). The top 18 art styles discovered by SDXL-SliderSpace are presented in Figure~\ref{fig:artsdxl}.

\subsection{Diversity Enhancement}
We provide additional qualitative examples demonstrating how our generic diversity sliders mitigate mode collapse in distilled models. Our observations indicate that distilled models such as DMD2~\cite{dmd} tend to generate visually similar images for identical prompts, despite different random seed initializations. Through our trained diversity sliders, which are model-agnostic, we successfully counter mode collapse (Section~\ref{sec:diverse_exp}). As quantitatively validated in Table~\ref{tab:diverse}, the diversity SliderSpace significantly improves image variation, achieving FID scores comparable to the base model.

\subsection{Concept Decomposition}
\begin{figure*}[!htbp]
    \centering
    \includegraphics[width=.8\linewidth]{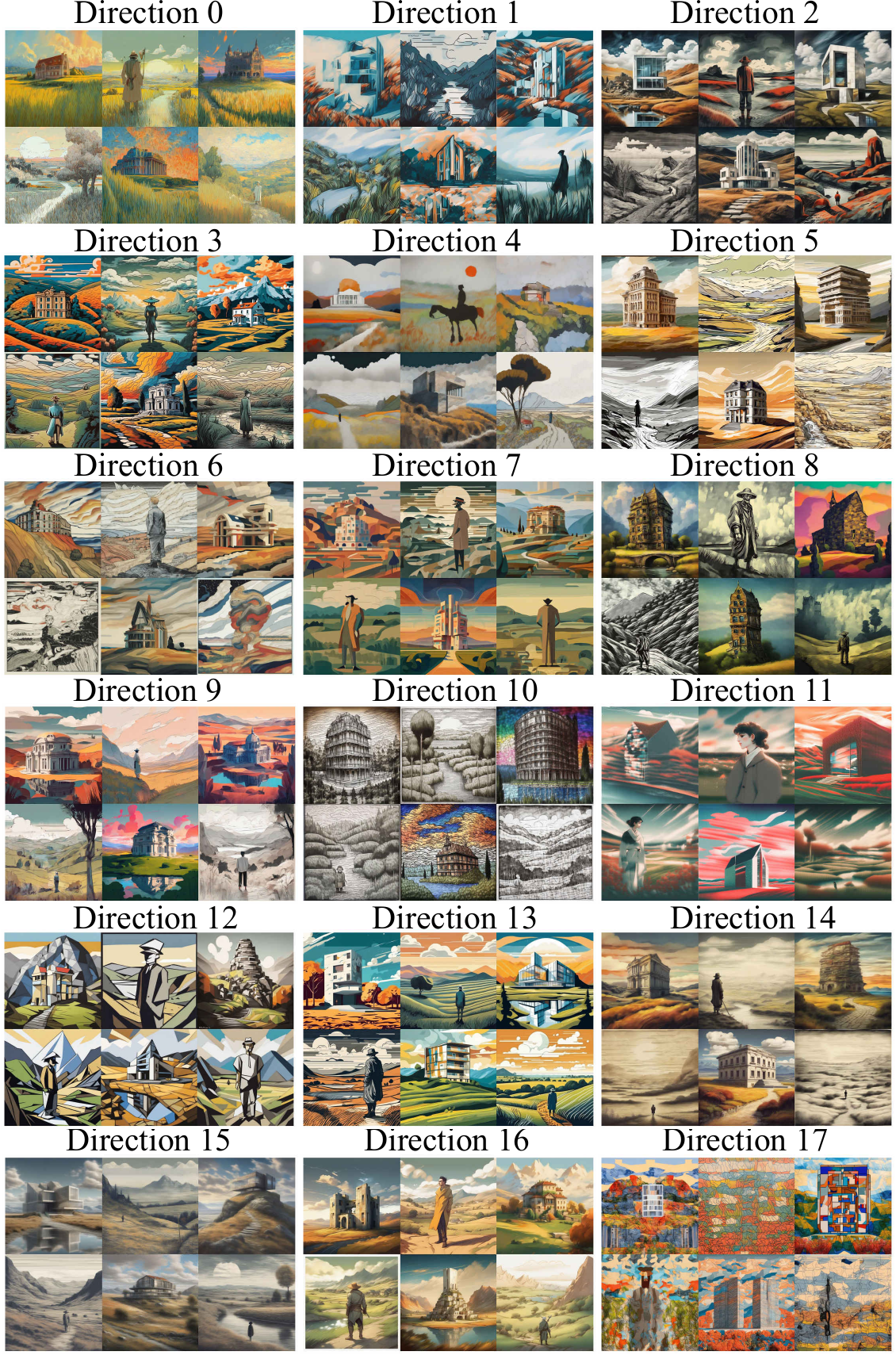}
    \caption{We show the top 18 art directions that are discovered in the SDXL~\cite{podell2023sdxl} SliderSpace for the concept ``art''.} 
    \label{fig:artsdxl}
\end{figure*}

\begin{figure*}
    \centering
    \includegraphics[width=\linewidth]{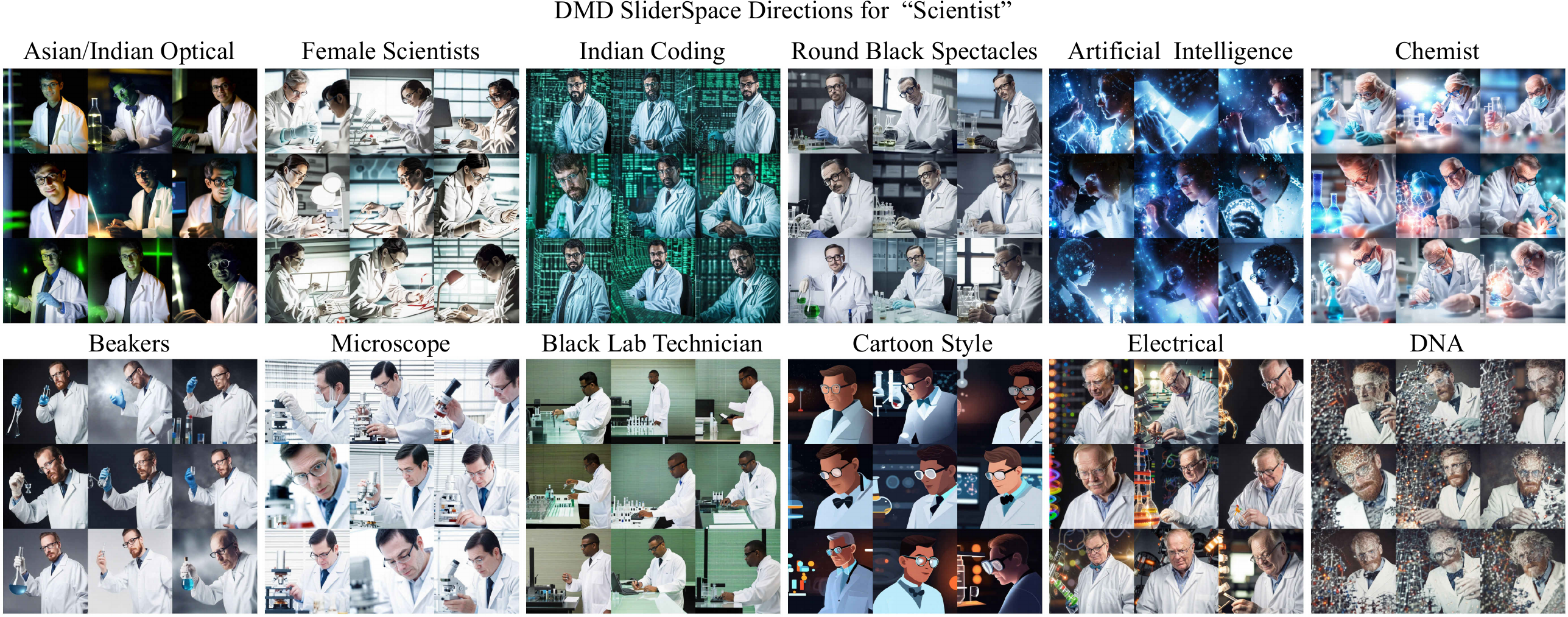}
    \caption{We show the SliderSpace discovered in SDXL-DMD2 4-step model~\cite{dmd} for the concept ``Scientist''} 
    \label{fig:concept1}
\end{figure*}

\begin{figure*}
    \centering
    \includegraphics[width=\linewidth]{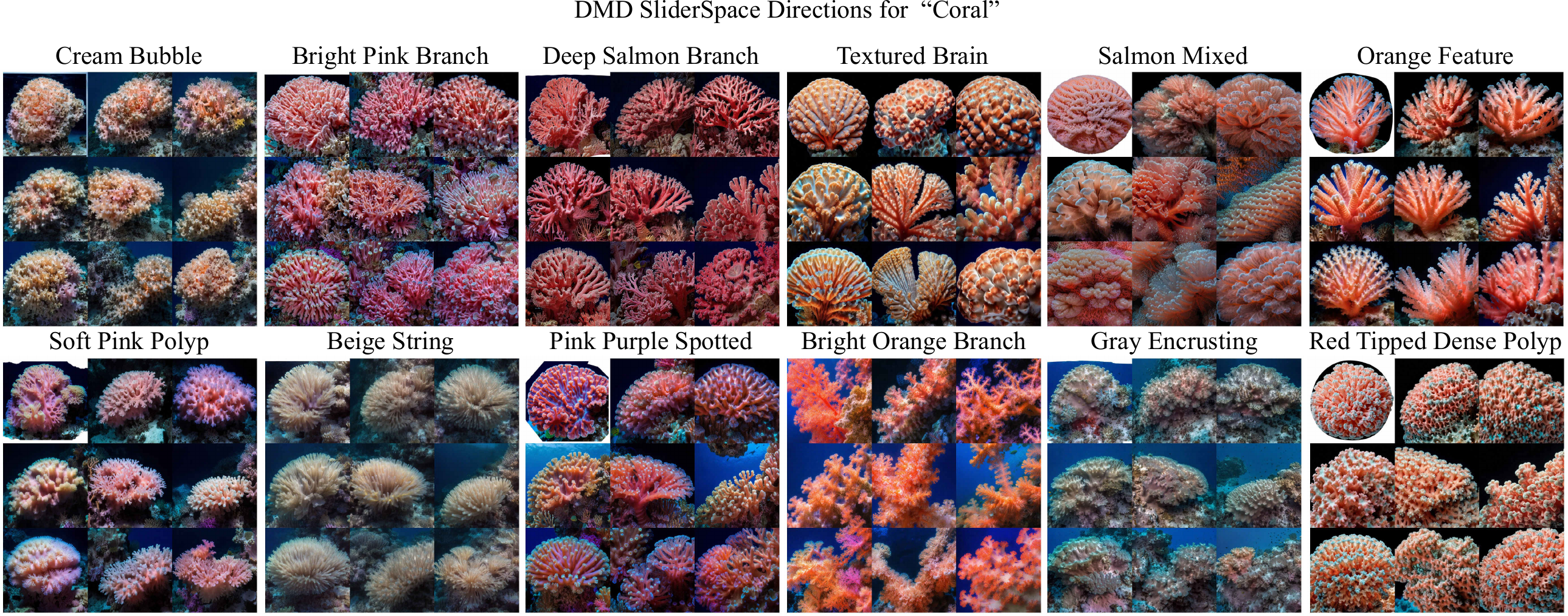}
    \caption{We show the SliderSpace discovered in SDXL-DMD2 4-step model~\cite{dmd} for the concept ``Coral''} 
    \label{fig:concept2}
\end{figure*}

\begin{figure*}
    \centering
    \includegraphics[width=\linewidth]{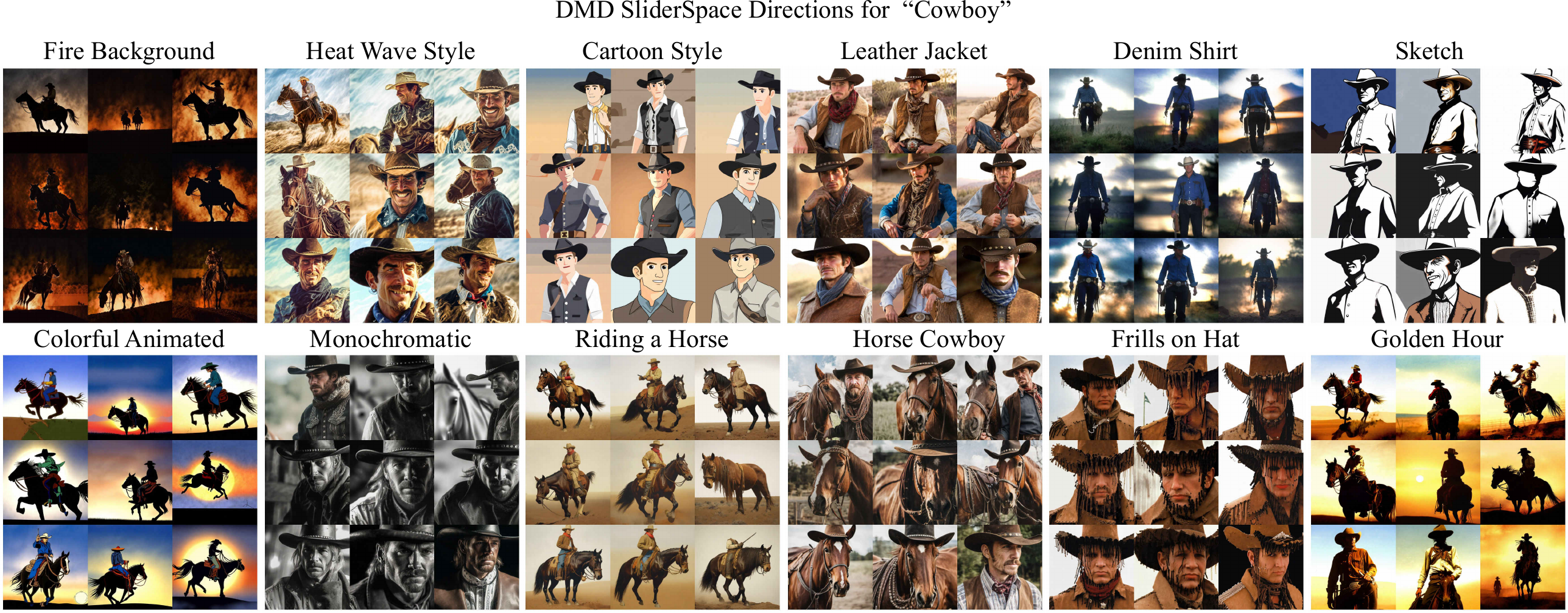}
    \caption{We show the SliderSpace discovered in SDXL-DMD2 4-step model~\cite{dmd} for the concept ``Cowboy''} 
    \label{fig:concept3}
\end{figure*}

\begin{figure*}
    \centering
    \includegraphics[width=\linewidth]{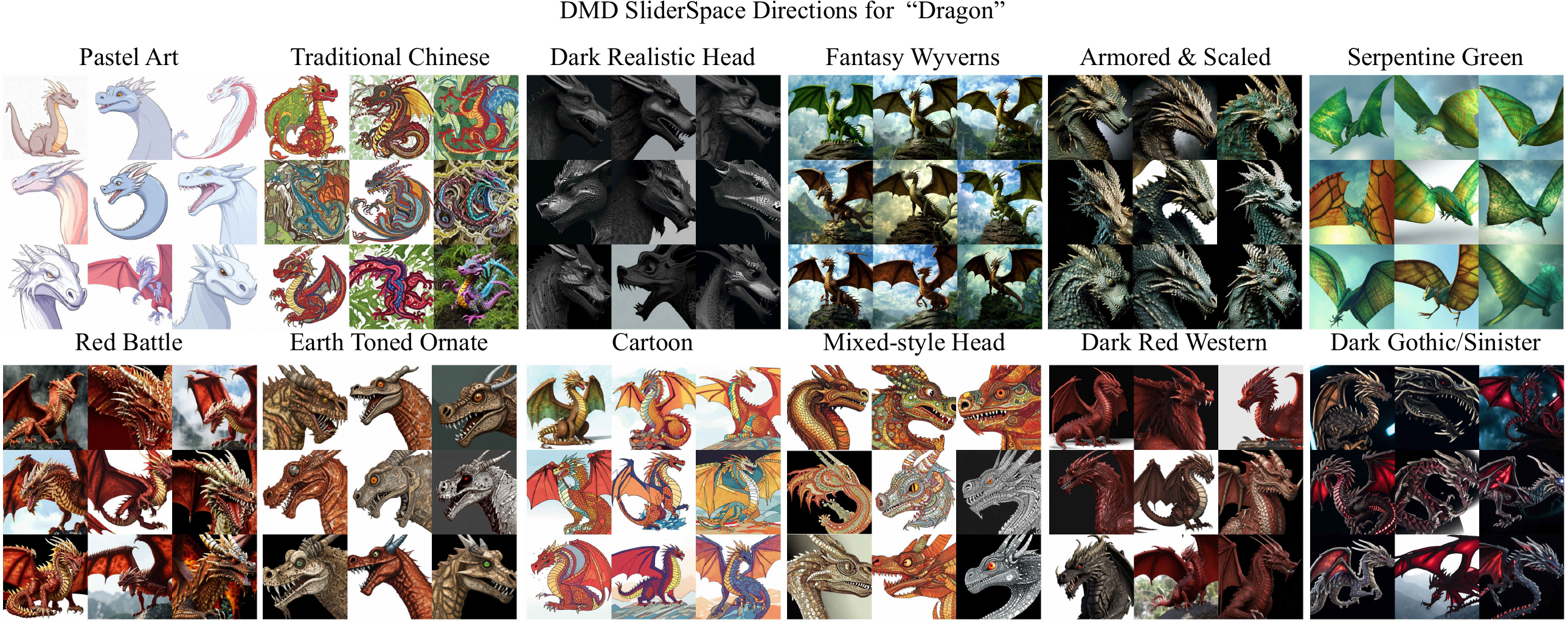}
    \caption{We show the SliderSpace discovered in SDXL-DMD2 4-step model~\cite{dmd} for the concept ``Dragon''} 
    \label{fig:concept4}
\end{figure*}

\begin{figure*}
    \centering
    \includegraphics[width=\linewidth]{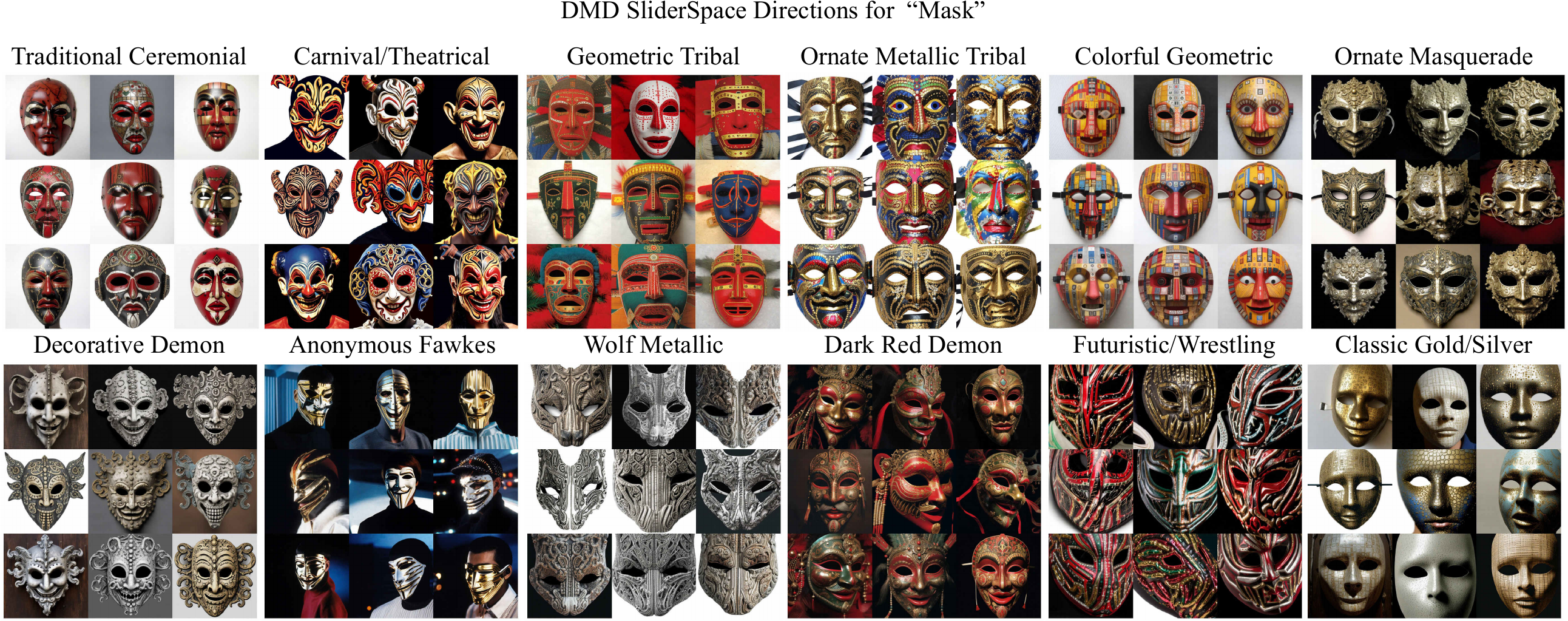}
    \caption{We show the SliderSpace discovered in SDXL-DMD2 4-step model~\cite{dmd} for the concept ``Mask''} 
    \label{fig:concept5}
\end{figure*}

\begin{figure*}
    \centering
    \includegraphics[width=\linewidth]{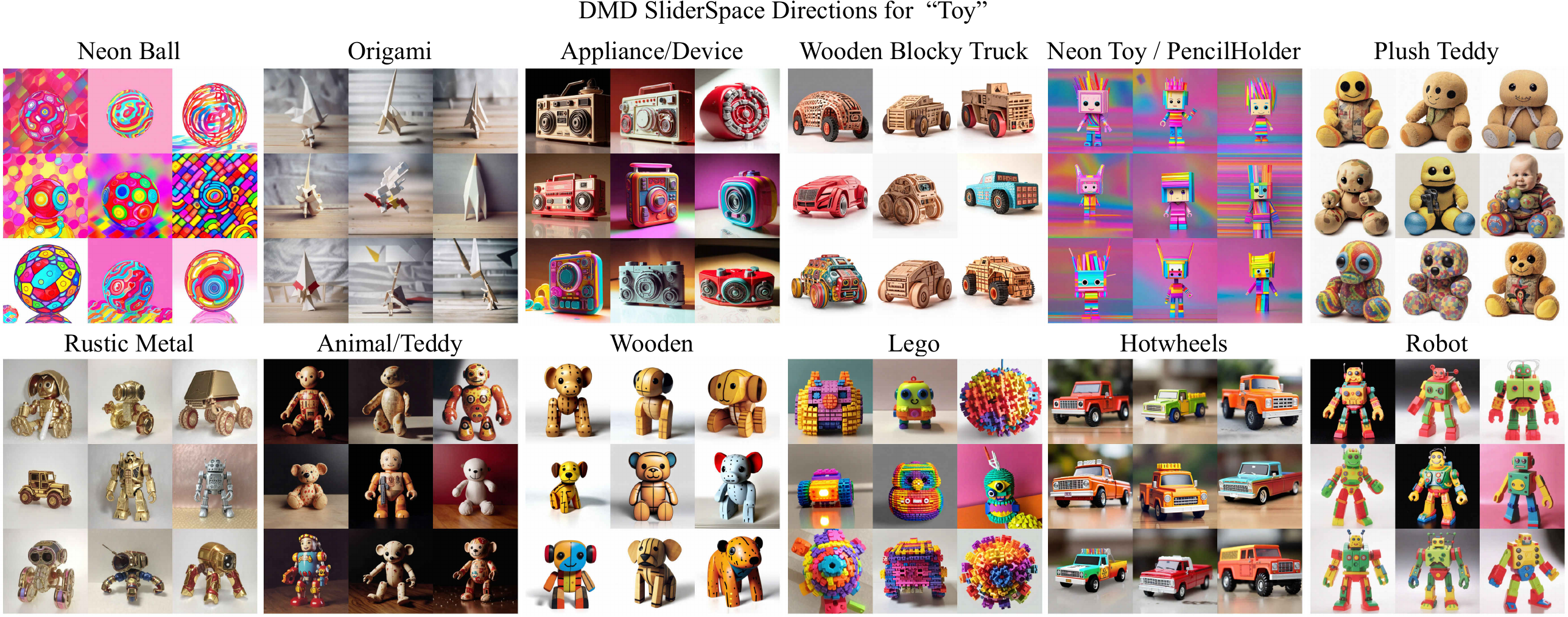}
    \caption{We show the SliderSpace discovered in SDXL-DMD2 4-step model~\cite{dmd} for the concept ``Toy''} 
    \label{fig:concept6}
\end{figure*}

\begin{figure*}
    \centering
    \includegraphics[width=\linewidth]{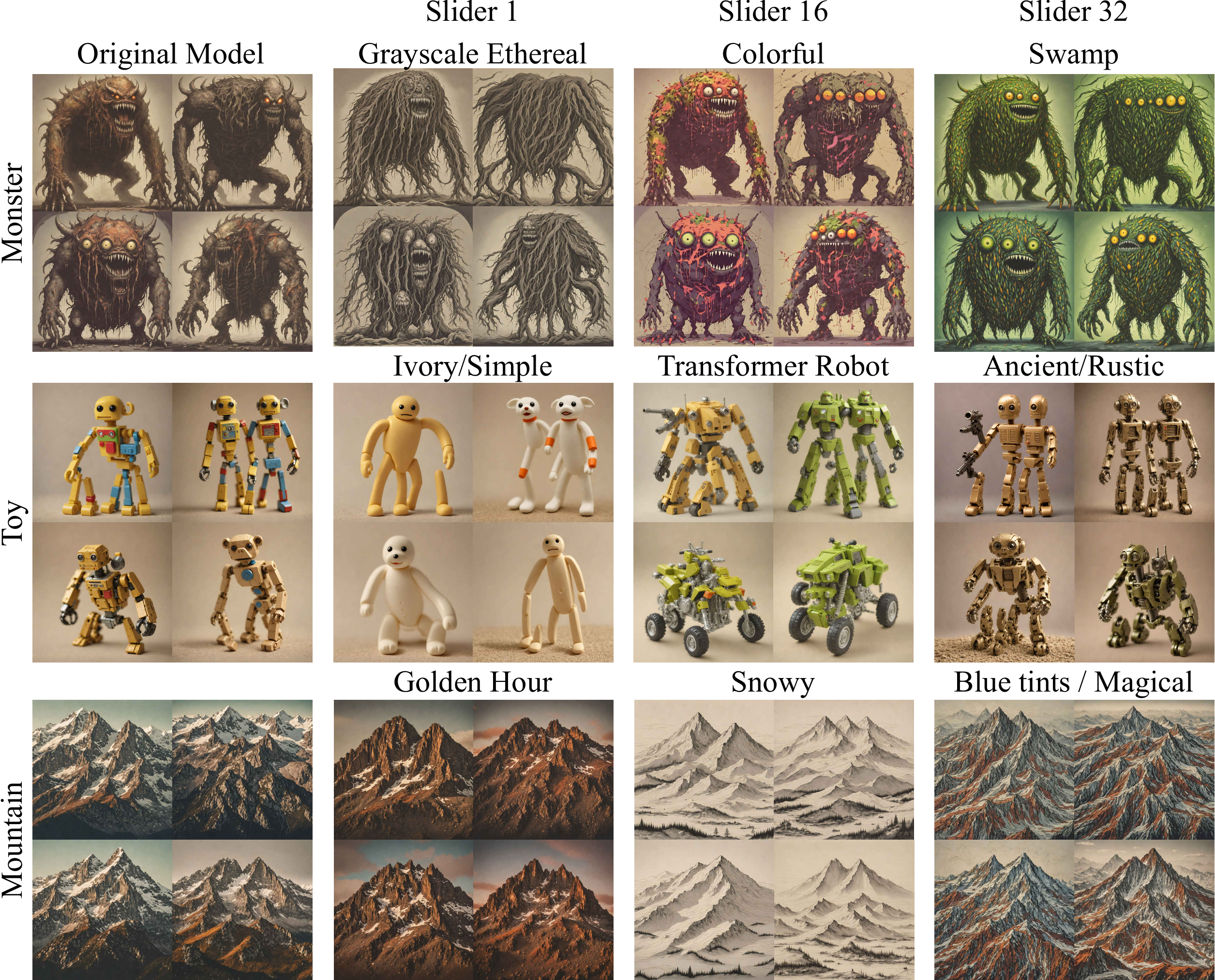}
    \caption{We show the SliderSpace discovered in SDXL-Turbo 4-step model~\cite{turbo} and how they can be used for precise control of image generation} 
    \label{fig:turbo1}
\end{figure*}

\begin{figure*}
    \centering
    \includegraphics[width=\linewidth]{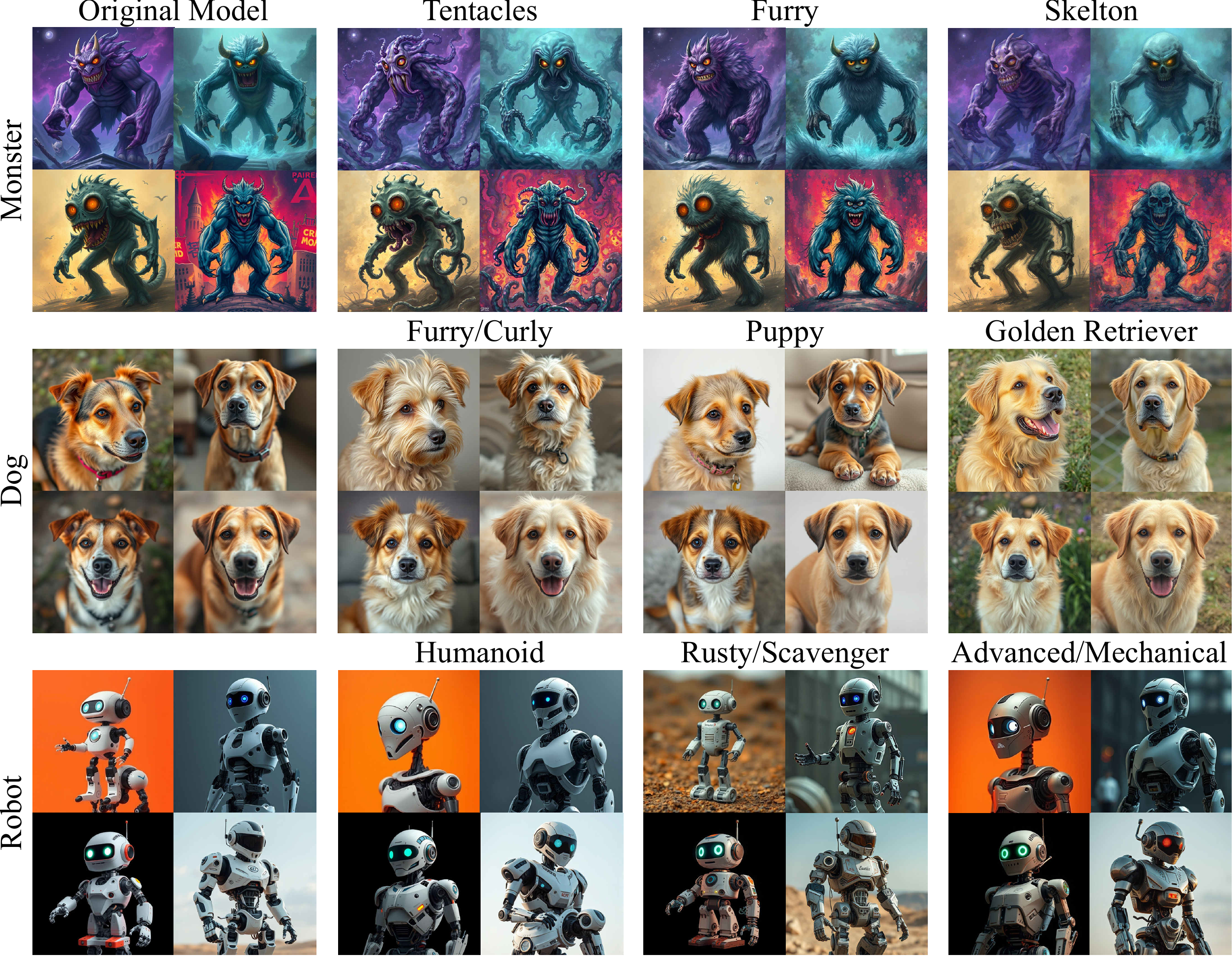}
    \caption{We show the SliderSpace discovered in FLUX Schnell model~\cite{flux} for concepts ``monster'' and ``dog''} 
    \label{fig:flux}
\end{figure*}

We present qualitative examples of concept decomposition using the SDXL-DMD2~\cite{dmd} SliderSpace in Figures~\ref{fig:concept1}--\ref{fig:concept6}. Furthermore, we demonstrate SliderSpace's versatility across various models, including SDXL-Turbo~\cite{turbo} (Figures~\ref{fig:turbo1}, SDXL-Base~\cite{podell2023sdxl} (Figures~\ref{fig:artsdxl}), and the state-of-the-art transformer-based FLUX Schnell models (Figures~\ref{fig:intro} and~\ref{fig:flux}). We note that Claude3.5~\cite{claude} generated captions are not always accurate. For instance, in Figure~\ref{fig:concept1}, Claude annotates one of the sliders as ``Black Lab Technician'', but it is not visually distinct whether the slider is `lab technician' or a `scientist'.

\section{Ablations}
\label{sec:ablations}
\begin{figure*}[!htp]
    \centering
    \includegraphics[width=\linewidth,trim=0 5mm 0 0]{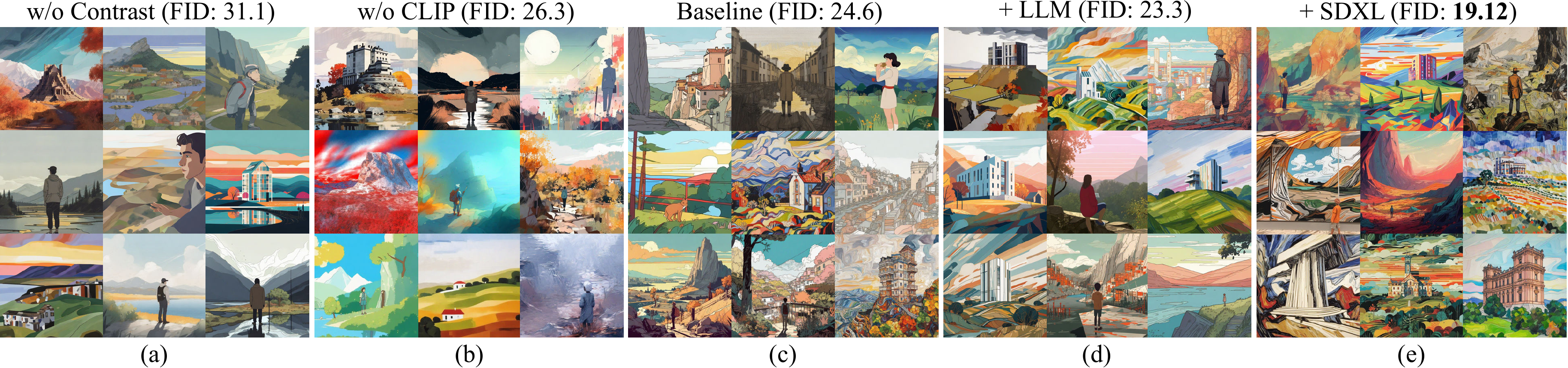}
    \caption{We conduct our ablations on the art-exploration application and show FID scores as a measure of diversity. SliderSpace contrastive objective (Eq.~\ref{eq:contrast} is essential for discovering diverse directions. Ablating CLIP space analysis and performing spectral analysis in diffusion output space (Eq.~\ref{eq:x0}) results in sliders that control color, texture and shapes. We also find that expanding the training data diversity using LLM enhanced prompts and base SDXL models can help with improved distilled model's SliderSpace diversity} 
    \label{fig:ablations}
\end{figure*}

We analyze the key components of our method and validate their necessity: (1) the semantic orthogonality objective, (2) expanding diversity of training samples, and (3) CLIP embedding analysis. Figure~\ref{fig:ablations} shows qualitative examples and FID measures on art exploration experiments. In both the qualitative and quantitative experiments, we find that uniqueness criteria in Eqn~\ref{eq:contrast} is very important to get diverse discovery of SliderSpace. When we extract a naive-SliderSpace by training multiple sliders on a single concept using regular customization~\cite{kumari2022customdiffusion,ruiz2022dreambooth} loss and no contrastive objective, many redundant and junk directions appear, as shown in Fig.~\ref{fig:ablations}(a). This baseline is equivalent to Liu et al.~\cite{liu2023unsupervised}. Similarly, by applying our objective (Eq.~\ref{eq:contrast}) on diffusion output space $\Tilde{x}_{0,t}$ (Eq.~\ref{eq:x0}) rather than CLIP space, SliderSpace discovers directions that are more relevant in color and shape but not semantic variations, as shown in Fig.~\ref{fig:ablations}(b). This baseline is slider equivalent version of NoiseCLR~\cite{dalva2024noiseclr}. Finally, diversity expansion of training data (Fig.~\ref{fig:ablations} d,e), helps with expanding a diverse set of sliders. This can be used to improve the variation across sliders. We use SDXL for generating images in concept and art experiments. For diversity experiments, we use LLM prompt expansion as we compare against SDXL as baseline.

\begin{figure*}
    \centering
    \includegraphics[width=\linewidth]{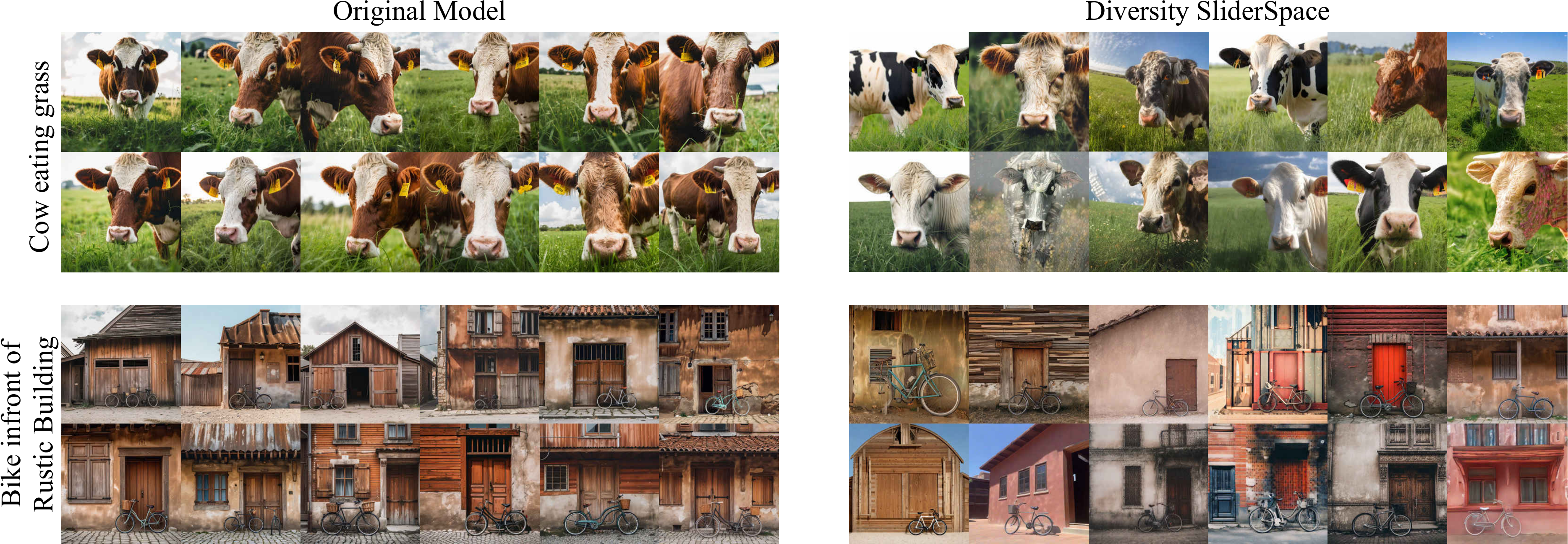}
    \caption{We show samples from our diversity experiments ~\ref{sec:diverse_exp}. We sample random 3 sliders from the SDXL-DMD2~\cite{dmd} diversity SliderSpace. We find that the common diversity sliderspace has a visual improvement in diversity and reverses the mode collapse in the distilled models}
    \label{fig:diverse1}
\end{figure*}

\begin{figure*}
    \centering
    \includegraphics[width=\linewidth]{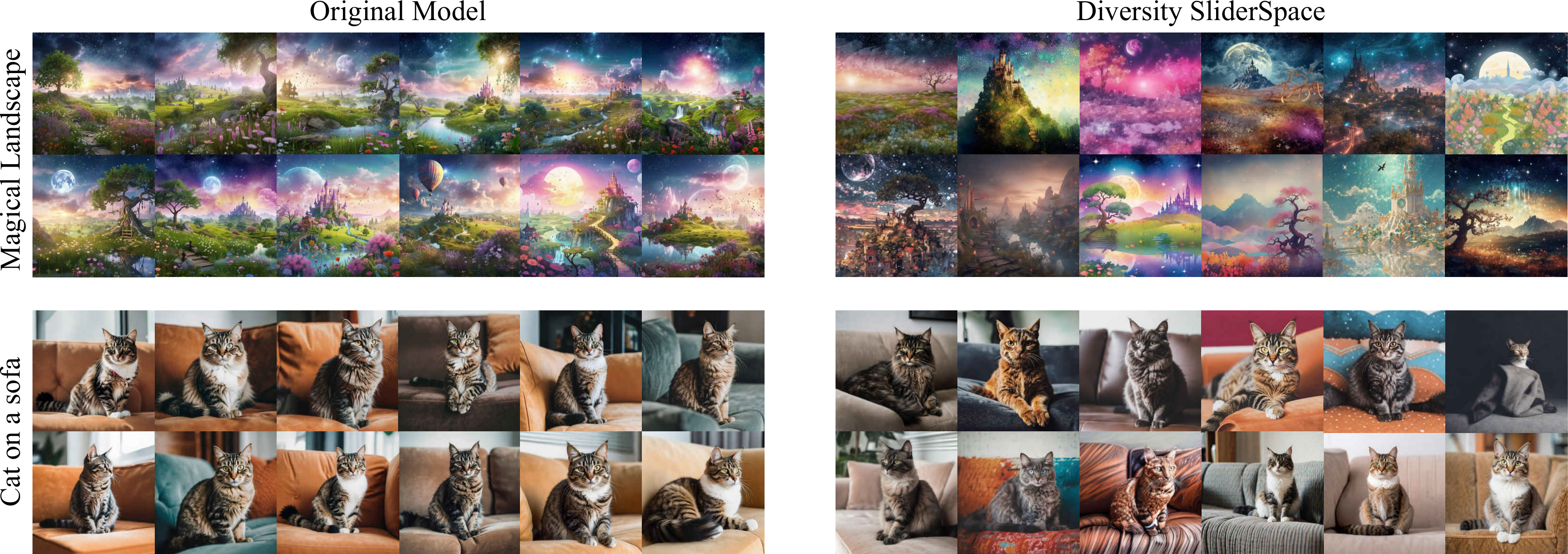}
    \caption{We show samples from our diversity experiments ~\ref{sec:diverse_exp}. We sample random 3 sliders from the SDXL-DMD2~\cite{dmd} diversity SliderSpace. We find that the common diversity sliderspace has a visual improvement in diversity and reverses the mode collapse in the distilled models}
    \label{fig:diverse2}
\end{figure*}

.

\end{document}